%% file: arxiv.tex
\documentclass[sigconf]{acmart}
\AtBeginDocument{%
  \providecommand\BibTeX{{%
    \normalfont B\kern-0.5em{\scshape i\kern-0.25em b}\kern-0.8em\TeX}}}

\usepackage{float}
\usepackage{newfloat}
\usepackage{multirow}
\usepackage{makecell}
\usepackage{xcolor}         % colors
\usepackage{colortbl}
\usepackage{dsfont}
\usepackage{amsmath}
\usepackage{enumitem}

\usepackage[normalem]{ulem}
\useunder{\uline}{\ul}{}

\definecolor{myyellow}{rgb}{1,1, 0.6}
\definecolor{myorange}{rgb}{1, 0.8, 0.6}
\definecolor{myred}{rgb}{1, 0.6, 0.6}
\definecolor{second}{HTML}{FFDAB9}
\definecolor{best}{HTML}{FFC1C1}
\usepackage{multirow}
\usepackage{colortbl}
\usepackage[normalem]{ulem}
\usepackage{bbding}
\useunder{\uline}{\ul}{}
\usepackage{stackrel}
\usepackage{todonotes}
\renewcommand\footnotetextcopyrightpermission[1]{} % 去掉页脚中的版权声明
\usepackage{enumitem}
\usepackage{algpseudocode}
\usepackage[ruled,lined]{algorithm2e}
\usepackage[normalem]{ulem}
\usepackage{tcolorbox}
\usepackage{balance}
\usepackage{arydshln}

\newcommand{\methodBrief}{Cas-SEAT}

\renewcommand{\todo}[1]{\iffalse #1 \fi{\color{blue} \textbf{[TODO]}}}

\newcommand{\leftrarrows}{\mathrel{\raise.9ex\hbox{\oalign{%
  $\scriptstyle\leftarrow$\cr
  \vrule width0pt height.5ex$\hfil\scriptstyle\relbar$\cr}}}}
\newcommand{\lrightarrows}{\mathrel{\raise.9ex\hbox{\oalign{%
  $\scriptstyle\relbar$\hfil\cr
  $\scriptstyle\vrule width0pt height.5ex\smash\rightarrow$\cr}}}}
\newcommand{\Rrelbar}{\mathrel{\raise.9ex\hbox{\oalign{%
  $\scriptstyle\relbar$\cr
  \vrule width0pt height.5ex$\scriptstyle\relbar$}}}}

\DeclareMathOperator*{\argminA}{arg\,min}

\begin{document}

% \title{Semantic Codebook Learning for Dynamic Recommendation Models}
% \title{Cascaded Self-Evaluation Augmented Training\\for Efficient Multimodal Large Language Models}
\title{Cascaded Self-Evaluation Augmented Training\\for Lightweight Multimodal LLMs}
\author{
    Zheqi Lv\textsuperscript{1}, Wenkai Wang\textsuperscript{1}, Jiawei Wang\textsuperscript{2}, Shengyu Zhang\textsuperscript{1}, Fei Wu\textsuperscript{1}\\
    \textsuperscript{1}Zhejiang University, Hangzhou, China \\
    \textsuperscript{2}National University of Singapore, Singapore \\
}

% \author{Zheqi Lv}
% \affiliation{%
%   \institution{Zhejiang University}
%   \city{Hangzhou}
% \email{zheqilv@zju.edu.cn}

% \author{Shaoxuan He}
% \affiliation{%
%   \institution{Zhejiang University}
%   \city{Hangzhou}
% \email{shxhe@zju.edu.cn}

% \author{Tianyu Zhan}
% \affiliation{%
%   \institution{Zhejiang University}
%   \city{Hangzhou}
% \email{yuzt@zju.edu.cn}

% \author{Shengyu Zhang}
% \affiliation{%
%   \institution{Zhejiang University}
%   \city{Hangzhou}
%   \country{China}\\
%   \institution{Shanghai Institute for Advanced Study, Zhejiang University}
%   \city{Shanghai}
% \authornote{Corresponding authors.}
% \email{sy_zhang@zju.edu.cn}

% \author{Wenqiao Zhang}
% \affiliation{%
%   \institution{Zhejiang University}
%   \city{Hangzhou}
% \email{wenqiaozhang@zju.edu.cn}

% \author{Jingyuan Chen}
% \affiliation{%
%   \institution{Zhejiang University}
%   % \streetaddress{1 Th{\o}rv{\"a}ld Circle}
%   \city{Hangzhou}
% \email{jingyuanchen@zju.edu.cn}

% \author{Zhou Zhao}
% \affiliation{%
%   \institution{Zhejiang University}
%   \city{Hangzhou}
% \email{zhaozhou@zju.edu.cn}

% \author{Fei Wu}
% \affiliation{%
%   \institution{Zhejiang University}
%   \city{Hangzhou}
% \email{wufei@zju.edu.cn}

\renewcommand{\shortauthors}{Zheqi Lv et al.}

\input{tex/0abstract}

%    <concept>
%        <concept_id>10002951.10003317.10003331.10003271</concept_id>
%        <concept_desc>Information systems~Personalization</concept_desc>
%        <concept_significance>500</concept_significance>
%        </concept>
%    <concept>
%        <concept_id>10002951.10003317.10003371.10003386</concept_id>
%        <concept_desc>Information systems~Multimedia and multimodal retrieval</concept_desc>
%        <concept_significance>500</concept_significance>
%        </concept>
%  </ccs2012>

% \ccsdesc[500]{Information systems~Personalization}
% \ccsdesc[500]{Information systems~Multimedia and multimodal retrieval}

% \keywords{Semenatic Codebook, Dynamic Model, Disentangle, Sequential Recommendation, Multimodal, Personalization}


\maketitle
\input{tex/1introduction}
\input{tex/2related_works}
\input{tex/3method}
\input{tex/4experiment}

\input{tex/5conclusion}
\clearpage
\bibliographystyle{ACM-Reference-Format}
\balance
\bibliography{reference}
\clearpage
\input{tex/9appendix}

\end{document}

%% file: tex/0abstract.tex
\begin{abstract}
\label{sec:abstract}

Efficient Multimodal Large Language Models (EMLLMs) can improve performance through Chain-of-Thought (CoT) reasoning, but they have poor self-evaluation capabilities during the CoT reasoning process. This is due to their tendency to simplify the reasoning process and the degradation of self-evaluation ability during downstream task fine-tuning. To address this, we intuitively propose \textit{Self-Evaluation Augmented Training (SEAT)}, which uses more powerful EMLLMs to evaluate CoT reasoning data. The evaluation data is then used to train EMLLMs. However, due to the difficulties EMLLMs face with processing long token input-output sequences, and the degradation of self-evaluation ability as a basis for CoT reasoning, the SEAT method is not fully adapted. Therefore, we further propose \textit{Cascaded Self-Evaluation Augmented Training (Cas-SEAT)}, which converts long prompts into cascaded short prompts, each focusing on a specific task. Additionally, we mix CoT reasoning and self-evaluation data to preserve its CoT reasoning ability while enhancing the self-evaluation capability of EMLLMs. We also conduct \textit{Double-level Data Filtering (DDF)}, which includes source data filtering and labeled data filtering, using both non-EMLLM and EMLLM for filtering. Cas-SEAT and DDF work together to improve the performance of EMLLMs. Experiments show that Cas-SEAT achieves an average improvement of 22.16\% across multiple datasets, and DDF significantly reduces the resource consumption of training~\footnote{The code and a portion of the Cas-SEAT dataset are available at \url{https://github.com/HelloZicky/Cas-SEAT}}.

\end{abstract}

%% file: tex/1introduction.tex
\section{Introduction}
\label{sec:introduction}

% \begin{figure*}[t]
%     \centering
%     \includegraphics[width=0.96\textwidth]{figure/introduction.pdf}
    
%     \vspace{-0.1cm}
%     \caption{Several modes of self-evaluation: (a) A multimodal dataset containing images, questions, and answers. (b) The typical approach in existing methods, where the model performs self-evaluation directly during inference. (c) The primary model synthesizes self-evaluation data and uses it for its own training.  (d) A more powerful model evaluates the primary model's outputs, and the primary model learns from this feedback to develop self-evaluation capabilities. (e) A more powerful model evaluates a small subset of the primary model's outputs, combines the evaluations with CoT data, and feeds them into the primary model. This approach also decomposes prompts and tasks to enable cascading self-evaluation capabilities.}
%     \label{fig:introduction}
%     % \vspace{+3mm}
%     % \vspace{-0.35cm}
%     \vspace{-0.3cm}
% \end{figure*}

\begin{figure*}[t]
    \centering
    \includegraphics[width=\textwidth]{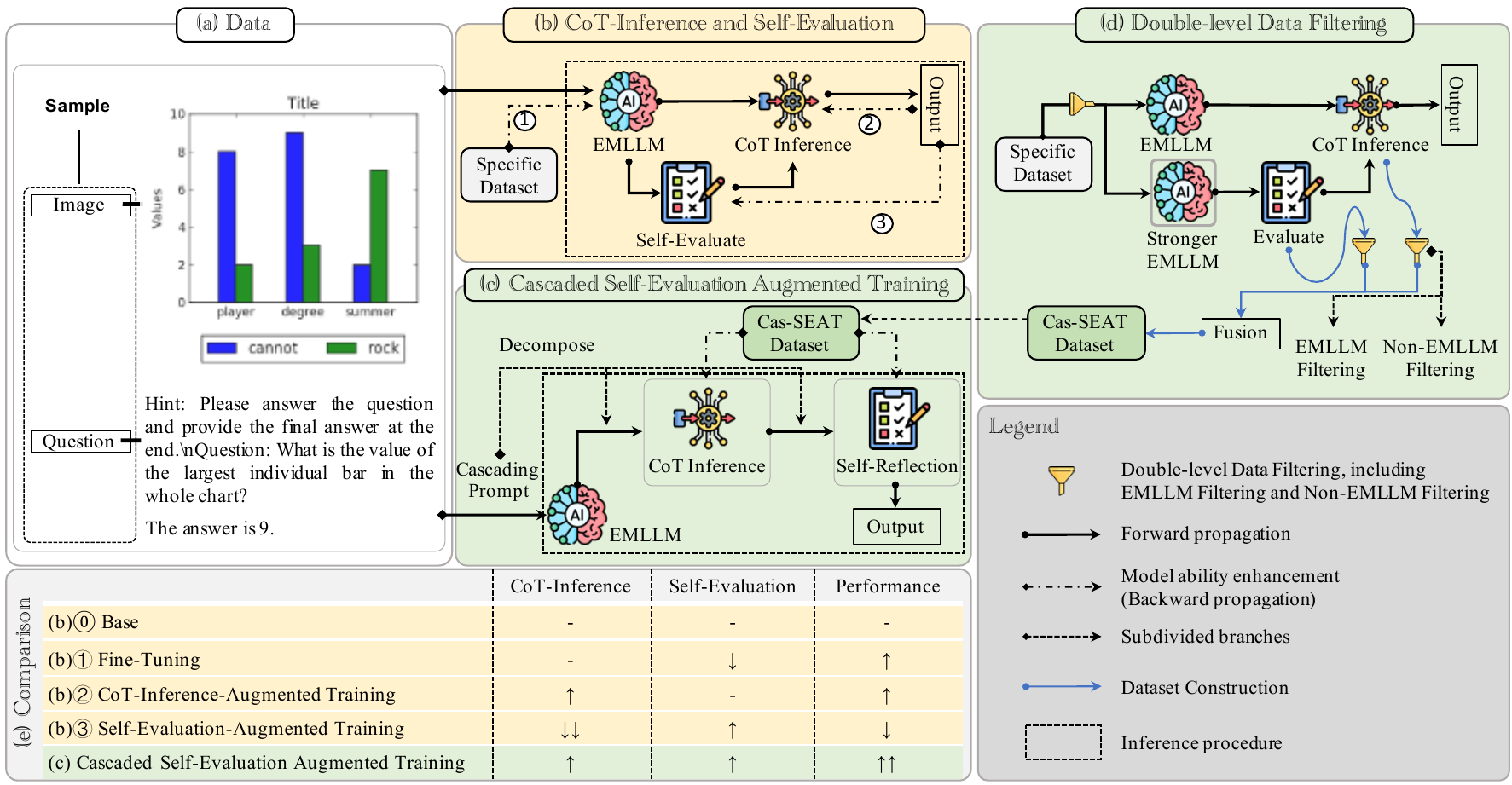}
    
    \vspace{-0.3cm}
    % \caption{Several modes of self-evaluation: (a) A multimodal dataset containing images, questions, and answers. (b) The typical approach in existing methods, where the model performs self-evaluation directly during inference. (c) The primary model synthesizes self-evaluation data and uses it for its own training.  (d) A more powerful model evaluates the primary model's outputs, and the primary model learns from this feedback to develop self-evaluation capabilities. (e) A more powerful model evaluates a small subset of the primary model's outputs, combines the evaluations with CoT data, and feeds them into the primary model. This approach also decomposes prompts and tasks to enable cascading self-evaluation capabilities.}
    \caption{
    % (a) A multimodal dataset containing images, questions, and answers. (b) Conventional CoT and self-evaluation reasoning and enhancement methods. (c) Our proposed computational method, Cas-SEAT. (d) Our proposed computational method
    (a) A sample from the dataset used for training and inference of multimodal large language models, containing images, questions, and answers. 
    (b) Overview of Chain-of-Thought (CoT) reasoning, self-evaluation reasoning, and their corresponding enhancement methods. 
    (c) The proposed computational method, Cas-SEAT. 
    (d) The proposed dataset construction method, DDF, which provides the Cas-SEAT Dataset for Cas-SEAT. 
    (e) Comparison of CoT reasoning ability, self-evaluation ability, and overall performance. Symbols ``–'', ``↑'', ``↓'', ``↑↑'', and ``↓↓'' indicate comparable, improved, degraded, significantly improved, and significantly degraded performance, respectively.
    }
    \label{fig:introduction}
    % \vspace{+3mm}
    % \vspace{-0.35cm}
    \vspace{-0.3cm}
\end{figure*}

In recent years, multimodal large language models (MLLMs) \cite{ref:llava1.5,ref:qwen2-vl} have developed rapidly. However, the growing demand for training and deployment in resource-constrained environments such as universities, hospitals, and communities has spurred interest in developing efficient multimodal large language models (EMLLMs) with smaller parameter sizes, such as 7B, 2B, or even 1B. Thanks to advances in model architecture, training methods, and techniques like chain-of-thought (CoT) reasoning \cite{ref:CoT}, EMLLMs can now generate step-by-step reasoning processes that are more logical and coherent.

With the widespread use of CoT reasoning, the accuracy of each step in the reasoning process has received increasing attention \citep{madaan2024self, zheng2023judging, liu2023llms}, which we refer to as self-evaluation. However, directly applying these techniques to EMLLMs has not been effective (as shown by the black bars in Figure~\ref{fig:method_bar}, where the improvement is minimal). This is primarily due to two reasons: (1) EMLLMs tend to ``cut corners'' when answering questions, making the reasoning process very short. (2) Due to the limited parameter size of EMLLMs, their self-evaluation ability is weak unless enhanced training is performed, rather than just pretraining. More critically, before deploying EMLLMs in applications, traditional supervised fine-tuning (SFT) is often applied to downstream tasks, which further weakens their already fragile self-evaluation ability (as shown in Table~\ref{tab:main_mathvista}, where fine-tuned EMLLMs perform worse in self-evaluation (Line 14) compared to pretrained models (Line 13)).

\begin{sloppypar}
To address these challenges, we intuitively propose a \textit{Self-Evaluation Augmented Training (SEAT)} method to enhance the self-evaluation capability of EMLLMs. SEAT leverages more powerful EMLLMs to perform CoT reasoning and evaluate them. By optimizing these evaluations, we generate high-quality reasoning samples that help improve the overall reasoning quality and self-evaluation ability of the target model (referred to as the ``main model'' in this paper). However, the vanilla SEAT method also faces two challenges: (1) Merging reasoning and evaluation tasks into a single prompt leads to excessively long inputs and outputs. From the input perspective, a long prompt is required for EMLLMs to handle both step-by-step reasoning and self-evaluation. From the output perspective, EMLLMs need to perform step-by-step reasoning followed by step-by-step self-evaluation. However, EMLLMs struggle with handling long token inputs and outputs. (2) The CoT reasoning ability of EMLLMs significantly drops after adding self-evaluation data for enhanced training, and CoT reasoning is the foundation of self-evaluation (as shown in the SEAT reasoning and evaluation performance in Tables~\ref{tab:main_general}, \ref{tab:main_mathvista}, \ref{tab:main_wemath}).
To overcome these limitations, we further improved SEAT by proposing the \textit{Cascaded Self-Evaluation Augmented Training (Cas-SEAT)} method. Cas-SEAT decouples the reasoning and self-evaluation processes into independent tasks, using a cascade of brief prompts, with each prompt focusing on a specific task. In order for Cas-SEAT to work, we primarily use labeled CoT data, mixed with a small amount of self-evaluation data. The mixed data is used for training the main model to ensure that the CoT reasoning ability is preserved while enhancing the self-evaluation capability of EMLLMs. Due to the weaker generalization ability of EMLLMs, many data beneficial for training large MLLMs are not suitable for EMLLMs, so we use Double-level Data Filtering (DDF) for data selection. DDF includes source data filtering and labeled data filtering, and the filtering methods involve both non-EMLLM and EMLLM filtering. DDF primarily considers selecting data beneficial for EMLLMs training from six aspects: image quality, text quality, text length, text format, problem domain, and problem difficulty. This ensures the model's responses are standardized while avoiding data that exceeds the capabilities of EMLLMs. DDF significantly reduces the amount of training data for Cas-SEAT, lowering its training cost and ensuring its applicability in resource-constrained environments. As the first work to explore self-evaluation augmented training, Cas-SEAT and the Cas-SEAT-DDF dataset we constructed provide a valuable foundation and reference for future research in this field.
\end{sloppypar}

Experimental results show that Cas-SEAT improves self-evaluation ability by 19.68\%, 55.57\%, and 46.79\% on the MathVista, Math-V, and We-Math datasets, respectively, significantly outperforming existing methods. Furthermore, with the combined effect of Cas-SEAT and DDF, we can use a 7B parameter open-source EMLLMs for data labeling, and the 7B model's performance exceeds that of a 13B model, fully demonstrating the potential of our method.

In summary, our contributions are as follows:

% \begin{itemize}[itemsep=-4pt,topsep=2pt]
\begin{itemize}
\item We analyzed the bottlenecks limiting the self-evaluation ability of EMLLMs and proposed the \textit{SEAT} method to enhance this ability. To our knowledge, this is the first research focused on self-evaluation for EMLLMs.

\item We designed \textit{Cas-SEAT}, which effectively addresses the challenges EMLLMs face when handling lengthy prompt inputs and long CoT and self-evaluation outputs. It successfully improves self-evaluation ability while maintaining CoT reasoning capability.

\item We designed \textit{DDF} and constructed the \textit{Cas-SEAT-DDF dataset}, the first dataset tailored for self-evaluation augmented training of EMLLMs. It is cost-effective and efficient, facilitating future research.

\item We conducted extensive experiments on multiple datasets, exploring the applicability of our method across different model architectures and parameter sizes. The results show that our method not only significantly improves the performance of EMLLMs but also outperforms many models that are much larger than the main model.

\end{itemize}

%% file: tex/2related_works.tex
\section{Related Work}
\label{sec:related_work}
% \paragraph{Reasoning Formulation.}
% Several studies have attempted to better formulate the reasoning problem. One approach is to generate rationales to enhance model interpretability~\citep{zhou2020interpretable, wiegreffe2021teach, wiegreffe2021measuring}. Recently, the focus has shifted towards decomposing the reasoning process into intermediate steps before reaching the final answer~\citep{wei2022chain, zhou2022least, gao2022pal, chen2022program}. Various decomposition techniques have been explored, such as question reduction~\citep{zhou2022least, yang2022seqzero}, iterative prompting~\citep{wang2022shepherd}, and chaining the steps~\citep{10.1145/3491102.3517582}. While incorporating intermediate reasoning steps has resulted in substantial performance improvements, errors or imperfections can accumulate, especially when the chains become longer~\citep{wu2016googles, guo2018long}.
% As such, we utilize LLM self-evaluation as a stepwise criterion to improve the chaining process.
% \paragraph{CoT Reasoning.} 
\textbf{CoT Reasoning.} 
% CoT reasoning~\cite{huang2022large,hoffmann2022training,chowdhery2022palm,wei2022emergent,zhang2022automatic,kojima2022large,wei2022chain} enhances the performance of LLMs by forcing them to perform step-by-step reasoning. Following \citet{wei2022chain}, numerous studies have shown that LLMs can perform CoT reasoning through zero-shot CoT prompting (\cite{kojima2022large}) and few-shot CoT prompting (\cite{wei2022chain,zhang2022automatic}). Improvements have been made in areas such as self-consistency \citep{wang2022self}, least-to-most prompting \citep{zhou2022least}, dynamic least-to-most prompting \citep{drozdov2022compositional}, bootstrapping training \citep{zelikman2022star}, and self-training \citep{huang2022large}. \citet{wang2022self, wang2022rationale} refined the original CoT prompting by marginalizing over diverse reasoning paths, while \citet{zelikman2022star, huang2022large} improved CoT prompting through bootstrapping on self-generated CoT prompts. \citet{li2022making} introduced a voting classifier to filter sampled CoT prompts before the final prediction. Additionally, prompt enhancement and selection \citep{shum2023automatic}, meta-heuristic approaches \citep{pan2023plum}, and meta-graph prompting \citep{pan2024pomp} have further advanced standard CoT prompting. 
Chain-of-Thought (CoT) reasoning~\cite{huang2022large,hoffmann2022training,chowdhery2022palm,wei2022emergent,zhang2022automatic,kojima2022large,wei2022chain} improves the performance of large language models (LLMs) by forcing them to perform step-by-step reasoning. Subsequent improvements in CoT reasoning for LLMs have been made in areas such as zero-shot CoT prompting~\cite{kojima2022large}, few-shot CoT prompting~\cite{wei2022chain,zhang2022automatic}, self-consistency~\citep{wang2022self}, multiple reasoning paths~\citet{wang2022self, wang2022rationale}, minimal-to-maximal prompting~\citep{zhou2022least}, dynamic minimal-to-maximal prompting~\citep{drozdov2022compositional}, guided training~\citep{zelikman2022star}, and self-training~\citep{zelikman2022star, huang2022large}. To optimize CoT prompts, \citet{li2022making} proposed filtering CoT prompts using a voting classifier before the final prediction, \citep{shum2023automatic} focused on prompt enhancement and selection, and meta-heuristic methods~\citep{pan2023plum} and meta-graph prompts~\citep{pan2024pomp} have also been employed to further optimize CoT prompting.
% CoT reasoning~\cite{huang2022large,hoffmann2022training,chowdhery2022palm,wei2022emergent,zhang2022automatic,kojima2022large} enhances the performance of LLMs by forcing them to perform step-by-step reasoning. The initial idea was proposed by \citet{wei2022chain}, demonstrating that simply modifying the prompt could significantly improve model performance on complex tasks. Following \citet{wei2022chain}, numerous studies have shown that LLMs can perform CoT reasoning through zero-shot CoT prompting (\cite{kojima2022large}) and few-shot CoT prompting (\cite{wei2022chain,zhang2022automatic}). Improvements have been made in areas such as self-consistency \citep{wang2022self}, least-to-most prompting \citep{zhou2022least}, dynamic least-to-most prompting \citep{drozdov2022compositional}, bootstrapping training \citep{zelikman2022star}, and self-training \citep{huang2022large}. \citet{wang2022self, wang2022rationale} refined the original CoT prompting by marginalizing over diverse reasoning paths, while \citet{zelikman2022star, huang2022large} improved CoT prompting through bootstrapping on self-generated CoT prompts. \citet{li2022making} introduced a voting classifier to filter sampled CoT prompts before the final prediction. Additionally, prompt enhancement and selection \citep{shum2023automatic}, meta-heuristic approaches \citep{pan2023plum}, and meta-graph prompting \citep{pan2024pomp} have further advanced standard CoT prompting. 
Existing CoT research paid little attention to correctness checking at each step, particularly lacking focus on EMLLMs. However, for EMLLMs, CoT reasoning is more prone to shortcuts and mistakes. Our Cas-SEAT improves the length and accuracy of CoT reasoning in EMLLMs through self-evaluation.
% Existing CoT studies pay little attention to the correctness of synthesized CoT data, especially lacking focus on the CoT capabilities of EMLLMs. CoT reasoning is more challenging for EMLLMs. Our Cas-SEAT improves the quality of synthesized training data through self-evaluation and enhances the CoT reasoning capabilities of EMLLMs.
% Although they significantly improved the ability of CoT inference, they pay relatively little attention to the correctness of the synthesized CoT data, especially regarding the CoT capabilities of EMLLMs. CoT reasoning is more challenging for EMLLMs with limited parameters, as they tend to generate very few tokens and often produce incoherent outputs. Our approach, Cas-SEAT, enhances the quality of synthetic data used for training through self-evaluation and improves the CoT reasoning capabilities of EMLLMs.

% \vspace{-0.3cm}
% \paragraph{LLM Self-Evaluation.}
\textbf{LLM Self-Evaluation.}
Using additional evaluators to assess the correctness of inference data has been proven effective~\citep{li2022advance, xu2024can,liu2023g}, but self-evaluation eliminates the need for additional annotations and evaluators~\citep{li2022making}, making it more efficient. Currently, LLMs demonstrate strong calibration capabilities, and more and more research focuses on using prompts to enable LLMs to perform self-evaluation~\citep{rae2021scaling, kadavath2022language, guo2017calibration, kadavath2022language, jiang2021can, kuhn2023semantic, fu2023gptscore, golovneva2022roscoe, zhang2023coder, shinn2023reflexion, madaan2024self, paul2023refiner}. 
Kocmi and Federmann~\cite{kocmi2023gemba} and Xu et al.~\cite{xu2023instructscore} designed methods to guide LLMs to generate more fine-grained corrective annotations. In addition, Koo et al.~\cite{koo2023benchmarking}, Zheng et al.~\cite{zheng2023judging}, Chang et al.~\cite{chang2023speak}, Deutsch et al.~\cite{deutsch2022limitations}, and Liu et al.~\cite{liu2023llms} explored issues of self-amplifying bias and fairness in LLMs. Notably, the scaling up of models plays a crucial role in improving calibration capabilities~\citep{rae2021scaling, wei2022emergent}. Compared to larger MLLMs, EMLLMs are fragile, with inherently weaker self-evaluation abilities and an inability to handle long token inputs and outputs. Moreover, the foundation of self-evaluation—the CoT reasoning ability—is also weak. Our Cas-SEAT focuses on these issues and significantly improves the self-evaluation ability of EMLLMs.

%% file: tex/3method.tex
\section{Methodology}
\label{sec:methodology}
\begin{figure*}
    \centering
    \includegraphics[width=1\textwidth,height=7.6cm]{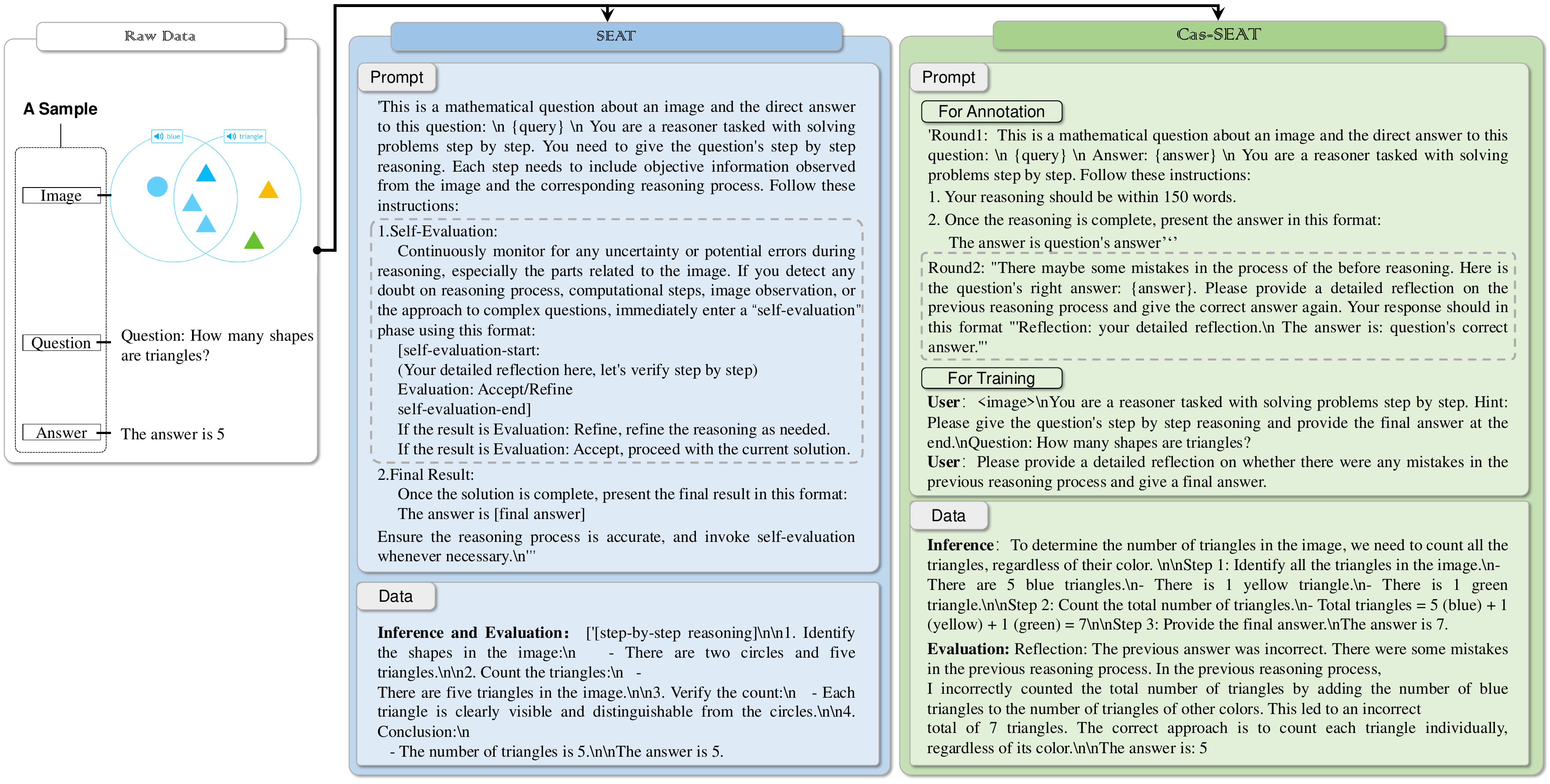}
    
    % \vspace{-0.1cm}
    \vspace{-0.3cm}
    \caption{
    Overview of the method. It illustrates the prompts designed for Augmented Self-Evaluation and Augmented Cascading Self-Evaluation, along with the corresponding training data generated.
    }
    \label{fig:method}
    % \vspace{+3mm}
    % \vspace{-0.35cm}
    \vspace{-0.3cm}
\end{figure*}

\subsection{Problem Formulation and Notations}
% \paragraph{Data and Model.} 
\subsubsection{Data and Model.} 
The primary EMLLM model is denoted as $\mathcal{M}$, with parameters $\Theta_0$. To achieve better performance on the target task, $\mathcal{M}$ needs to be fine-tuned on the dataset $\mathcal{D}$, 
We use $\{\mathcal{X}, \mathcal{Y}\}$ to represent a sample, $\mathcal{X}$ include images $\mathcal{I}$ and queries $\mathcal{Q}$, and $\mathcal{Y}$ represents answers. 
% We use $\{x, y\}$ to represent a sample.
After fine-tuning, $\Theta_0$ will be optimized to $\Theta$. The output of $\mathcal{M}$ are denoted as $\hat{\mathcal{Y}}$. $\mathcal{D}_{\rm{se}}$ represents the self-evaluation data generated by $\mathcal{M}$.
A more powerful EMLLM model is denoted as $\mathcal{G}$, which generates CoT reasoning data based on $\mathcal{D}$, denoted as $\mathcal{S}_{\rm{cot}}$. After evaluating $\mathcal{S}_{\rm{cot}}$, $\mathcal{G}$ produces self-evaluation training data, denoted as $\mathcal{S}_{\rm{se}}$. A small subset of $\mathcal{S}_{\rm{cot}}$ is selected, denoted as $\mathcal{S}_{\rm{cot}}^{\rm{sub}}$, and after being evaluated by $\mathcal{G}$, generates cascaded self-evaluation training data, denoted as $\mathcal{S}_{\rm{cse}}$. 

% \paragraph{Prompt.} 
\subsubsection{Prompt.}
The prompts used for $\mathcal{M}$ inference and self-evaluation are denoted as $P$ and $P_{se}$, respectively. The prompts used by $\mathcal{G}$ to generate cascading CoT data and evaluation data are denoted as $P_{\rm{cot}}$ and $P_{se}$, respectively. The prompts used by $\mathcal{G}$ to generate CoT data and evaluation data are denoted as $P_{\rm{cot}}$ and $P_{\rm{{cse}}}$, respectively. Identical symbols indicate identical prompts.

% \paragraph{Formula.} 
\subsubsection{Formula.}
% We formalize the existing studies to facilitate a formula-based comparison of their differences. $l$ represents the loss function.
We use $l$ to represent the loss function and $\mathcal{L}$ to represent the total loss.

\noindent\textit{Inference}:
\begin{equation}
\resizebox{0.38\textwidth}{!}{
$
\hat{\mathcal{Y}} = \mathcal{M}(\{P \!\!\quad \text{or} \!\!\quad P_{\rm{cot}} \!\!\quad \text{or} \!\!\quad P_{\rm{se}}\}, \mathcal{X}; \Theta_0), \mathcal{X} \in \mathcal{D}
$}
\end{equation}

\noindent\textit{Finetune}:
\begin{equation}
\resizebox{0.38\textwidth}{!}{
$
\argminA\limits_{\Theta_0} \mathcal{L} = \sum\limits_{\mathcal{X}, \mathcal{Y} \in \mathcal{D}} l(\mathcal{Y},\mathcal{M}(P \!\!\quad \text{or} \!\!\quad P_{\rm{cot}}, \mathcal{X}; \Theta_0))
% \mathcal{Y} = \mathcal{M}({P \text{or} P_{\rm{cot}}, \mathcal{X}; \Theta})
$}
\end{equation}

% \noindent\textit{Augmented CoT}:
% \begin{equation}
% \resizebox{0.4\textwidth}{!}{
% $
% \left\{
% \begin{aligned}
% &\mathcal{S}_{\rm{cot}} = \mathcal{G}(P_{\rm{cot}}, \mathcal{X}; \Theta) \\
% &\argminA_{\Theta} \mathcal{L} = \sum_{\mathcal{X}, \mathcal{Y}\in \mathcal{S}_{\rm{cot}}} l(\mathcal{Y},\mathcal{M}(P \!\!\quad \text{or} \!\!\quad P_{\rm{cot}}, \mathcal{X}; \Theta))
% \end{aligned}
% \right.
% $}
% \end{equation}

% \noindent\textit{Self-Evaluation}:
% \begin{equation}
% \mathcal{Y} = \mathcal{M}({P \!\!\quad \text{or} \!\!\quad P_{\rm{cot}}, \mathcal{X}; \Theta})
% \end{equation}

\noindent\textit{SEAT}:
\begin{equation}
\resizebox{0.35\textwidth}{!}{
$
\left\{
\begin{aligned}
&\mathcal{D}_{\rm{se}} = \mathcal{M}(P_{\rm{se}}, \mathcal{X}; \Theta) \\
% &\argminA_{\Theta} \mathcal{L} = \sum_{\{\mathcal{D},\mathcal{D}_{\rm{se}}\}} l(\mathcal{Y},\mathcal{M}(P_{\rm{se}}, \mathcal{X}; \Theta))
&\argminA_{\Theta} \mathcal{L} = \sum_{\mathcal{X}, \mathcal{Y}\in \mathcal{D}_{\rm{se}}} l(\mathcal{Y},\mathcal{M}(P_{\rm{se}}, \mathcal{X}; \Theta))
\end{aligned}
\right.
$
}
\label{eq:seat}
\end{equation}

\noindent\textit{Cas-SEAT}:
\begin{equation}
\resizebox{0.39\textwidth}{!}{
$
\left\{
\begin{aligned}
&\mathcal{S}_{\rm{cot}} = \mathcal{G}(P_{\rm{cot}}, \mathcal{X}; \Theta) \\
& \mathcal{S}_{\rm{cse}} = \mathcal{G}(P_{\rm{cse}}, \mathcal{X}; \Theta) \\
& \argminA_{\Theta} \mathcal{L} = \sum_{\mathcal{X}, \mathcal{Y}\in \{\mathcal{S}_{\rm{cot}},\mathcal{S}_{\rm{se}}\}} l(\mathcal{Y},\mathcal{M}(P_{\rm{cse}}, \mathcal{X}; \Theta))
\end{aligned}
\right.
\label{eq:cas-seat}
$}
\end{equation}

\subsection{Double-level Data Filtering}

Our data filtering process is conducted in two stages: source data filtering and labeled data filtering. The filtering methods include EMLLM filtering and non-EMLLM filtering. Samples that are filtered out are collectively referred to as ``rejected samples'' in this paper. In the \textit{source data filtering stage}, we use the MathV360K dataset~\cite{mathllava_math360k}, a commonly used dataset for multimodal large language model training, and apply EMLLM filtering. The filtering criteria are as follows: 1) Image quality: Based on image clarity. Samples with low image clarity are discarded. 2) Text quality: Based on the relevance between text and image and whether the question is clear and well-defined. Samples with mismatched text and images, or with vague questions, are discarded. 3) Question domain: Based on whether specialized domain knowledge is required. Overly specialized scientific questions, such as medical CT image diagnostics, are removed. 4) Question difficulty: Based on whether a more powerful model can answer it correctly. We randomly sample some samples and use Qwen2-VL-7B to attempt answers. If the accuracy of the answers is close to random values, the sample is discarded. In the \textit{labeled data filtering stage}, we use Qwen2-VL-7B to label the data, followed by non-EMLLM filtering based on Qwen's responses. The filtering criteria are as follows: 1) Text quality: Samples with garbled responses or responses in languages other than English are discarded. 2) Text length: Samples with excessively long responses are discarded. 3) Text format: The response format is standardized, and samples that do not conform to the format are discarded.

% \subsection{Augmented Self-Evaluation}
\subsection{Vanilla SEAT}
\begin{figure}[!ht]
    \centering
    \includegraphics[width=0.42\textwidth]{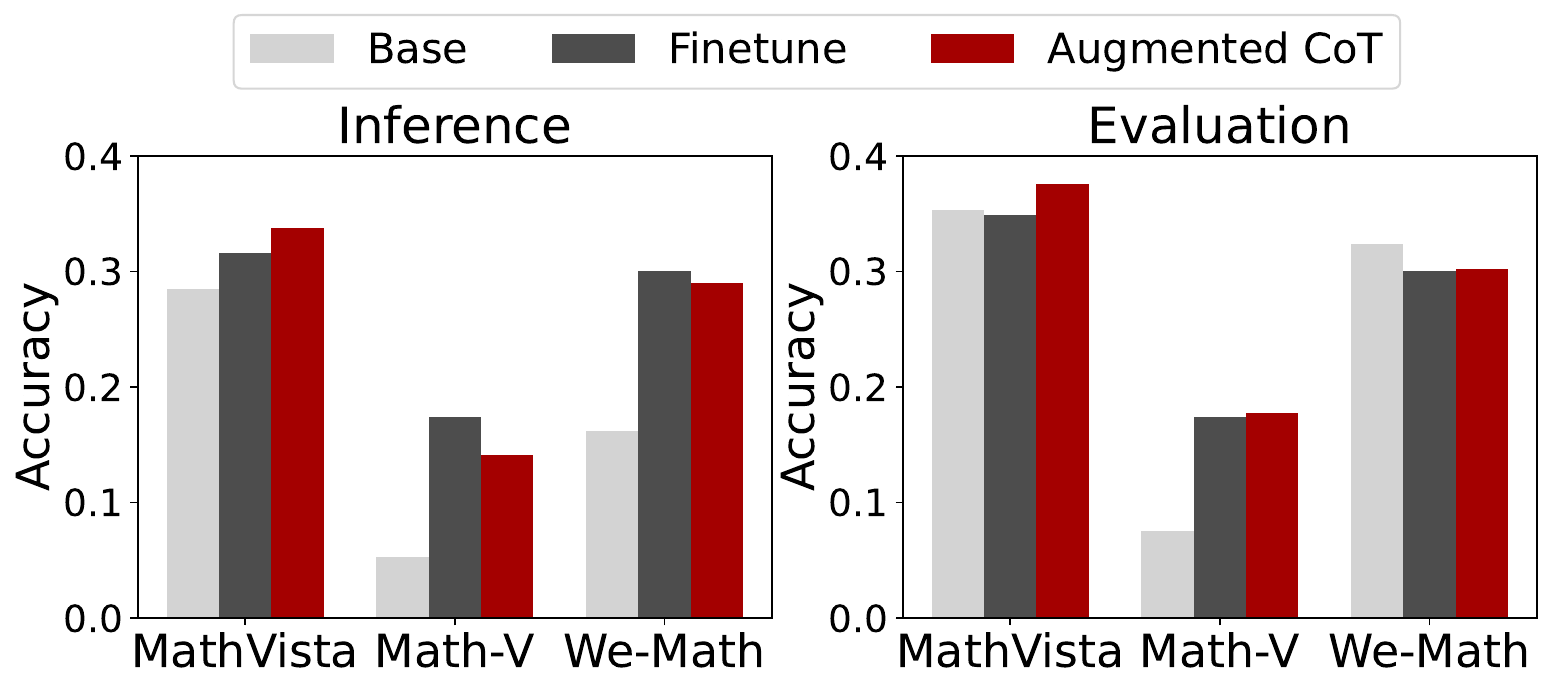}
    % \vspace{-0.1cm}
    \vspace{-0.3cm}
    \caption{
    Comparison of evaluation in terms of model performance improvement
    }
    \label{fig:method_bar}
    % \vspace{+3mm}
    \vspace{-0.7cm}
    
\end{figure}
Since self-evaluation has been proven effective in LLMs and MLLMs, we tested the self-evaluation performance of EMLLM, as shown in Figure~\ref{fig:method_bar}. Specifically, we evaluated inference and post-inference evaluation using the LLaVA-based Base(pretrained model), Finetune (finetune on Math360k), and Augmented CoT (Qwen2-VL (7B) annotates CoT data on Math360k, which is then learned by LLaVA-1.5.) on MathVista, Math-V, and We-Math. We observed that, in most cases, evaluation led to some improvement, though the improvement was not significant. However, in certain cases, evaluation resulted in a performance decline. We believe this variability in performance is primarily due to the lack of evaluation-specific data during training. Therefore, we aim to enhance the evaluation capabilities of EMLLM by synthesizing more self-evaluation data for its training.

The prompt \(P_{\rm{se}}\) and the training data \(\mathcal{S}_{\rm{se}}\) for Augmented Self-Evaluation are shown in Figure~\ref{fig:method}. The prompt divides Augmented Self-Evaluation into two parts: CoT reasoning and self-evaluation. The self-evaluation also includes data selection, where the MLLM autonomously determines which data to use for learning and which to discard.
The learning process can be formulated as Equation~\ref{eq:seat}.

\subsection{Cas-SEAT}
Due to the parameter limitations of EMLLM, its performance significantly degrades in scenarios involving long-token inputs and outputs. Consequently, Augmented Self-Evaluation often leads to a decline in performance. To address this issue, we decouple the inference and evaluation processes, as illustrated in Figure~\ref{fig:method}. First, we use a more powerful EMLLM $\mathcal{G}$ to obtain $\mathcal{S}_{\rm{cot}}$ based on $P_{\rm{cot}}$. Next, we continue to use $\mathcal{G}$ to evaluate the incorrect parts of $\mathcal{S}_{\rm{cot}}$, denoted as $Error(\mathcal{S}_{\rm{cot}})$. This process produces $\mathcal{S}_{\rm{cse}} = \mathcal{G}(P_{\rm{cse}}, Error(\mathcal{S}_{\rm{cot}}))$. Subsequently, we retain the data corrected through evaluation, denoted as $Correct(\mathcal{S}_{\rm{cse}})$, which satisfies the relationship $Correct(\mathcal{S}_{\rm{cse}}) \subset \mathcal{S}_{\rm{cse}}$. For simplicity, $Correct(\mathcal{S}_{\rm{cse}})$ is abbreviated as $\mathcal{S}_{\rm{cse}}$. 
So the whole equation to get $\mathcal{S}_{\rm{cse}}$ is:
\begin{equation}
\resizebox{0.4\textwidth}{!}{
$
% \left\{
\begin{aligned}
% &\mathcal{S}_{\rm{cot}} = \mathcal{G}(P_{\rm{cot}}, \mathcal{X}; \Theta) \\
& \mathcal{S}_{\rm{cse}} := Correct(\mathcal{G}(P_{\rm{cse}}, Error(\mathcal{G}(P_{\rm{cot}})))) \\
\end{aligned}
% \right.
\label{eq:cas-seat2}
$
}
\end{equation}

In the above equation, $A:=B$ means assigning the value of $B$ to $A$. Then, $\mathcal{M}$ is fine-tuned on $\mathcal{S}_{\rm{cse}}$ like Equation~\ref{eq:cas-seat}.

\input{table/main_table_general}

\input{table/main_table_mathvista}

%% file: table/main_table_general.tex
% \usepackage{multirow}
\begin{table*}[!ht]
\centering
\renewcommand{\arraystretch}{0.85}
\caption{Comparison of Cas-SEAT and the Baselines on MMMU Dataset. BUS, TE, AD, HM, SCI and HSS respectively denotes the Business, Tech and Engineering, Art and Design, Health and Medicine, Science, Humanities and Social Science.}
\label{tab:main_general}
\vspace{-0.3cm}
\resizebox{0.84\textwidth}{!}{
\begin{tabular}{ccccccccc}
\toprule[1.5pt]
 &  & \multicolumn{7}{c}{\textbf{MMMU}} \\ \cline{3-9}
% \multirow{-2}{*}{\textbf{Method}} & \multirow{-2}{*}{\textbf{Extra Data}} & Total Accuracy & Business & Tech and Engineering & Art and Design & Health and Medicine & Science & Humanities and Social Science \\
\multirow{-2}{*}{\textbf{Model}} & \multirow{-2}{*}{\textbf{Extra Data}} & Total Accuracy & BUS & TE & AD & HM & SCI & HSS \\
% \toprule \toprule
\midrule
\multicolumn{9}{c}{\textit{Heuristics Baselines}} \\
\midrule
Random Choice & Base & 0.2210 & 0.2470 & 0.2140 & 0.2920 & 0.2070 & 0.1800 & 0.2000 \\
\midrule
\multicolumn{9}{c}{\textit{Open-Source Models   Inference}} \\
\midrule
miniGPT4-7B & Base & 0.2680 & 0.2130 & 0.2380 & 0.2920 & 0.3070 & 0.2870 & 0.2920 \\
CogVLM-17B & Base & 0.3210 & 0.2560 & 0.2890 & 0.3800 & 0.3120 & 0.2510 & 0.4150 \\
LLaVA-v1.5-13B & Base & 0.3640 & 0.2270 & 0.3140 & 0.5170 & 0.3870 & 0.2930 & 0.5330 \\
% \hdashline{1-3}
\cdashline{1-9} % 贯穿第1列到第3列的横虚线
% \multicolumn{3}{c}{\hdashline} \\
\cellcolor[HTML]{F2F2F2} & Base & 0.3140 & 0.2133 & 0.2762 & 0.4667 & 0.2600 & 0.2200 & 0.5000 \\
\cellcolor[HTML]{F2F2F2} & Finetune & 0.3559 & 0.2600 & 0.3571 & 0.5133 & 0.3067 & 0.2800 & 0.4333 \\
\cellcolor[HTML]{F2F2F2} & CoT & 0.3118 & 0.1800 & 0.2666 & 0.4200 & 0.2800 & 0.2733 & 0.4916 \\
\cellcolor[HTML]{F2F2F2} & SEAT & 0.3204 & 0.2400 & 0.2857 & 0.4066 & 0.2800 & 0.2600 & 0.5000 \\
\multirow{-5}{*}{\cellcolor[HTML]{F2F2F2}LLaVA-v1.5-7B} & Cas-SEAT & 0.3226 & 0.2667 & 0.3238 & 0.3733 & 0.3400 & 0.2267 & 0.4250 \\
\midrule
\multicolumn{9}{c}{\textit{Open-Source Models   Self-Evaluation}} \\
\midrule
\cellcolor[HTML]{F2F2F2} & Base & 0.3677 & {\ul 0.3333} & 0.3286 & 0.4533 & 0.3467 & 0.2667 & {\ul 0.5250} \\
\cellcolor[HTML]{F2F2F2} & Finetune & {\ul 0.3946} & 0.2933 & {\ul 0.3667} & {\ul 0.5667} & {\ul 0.3933} & {\ul 0.3067} & 0.4667 \\
\cellcolor[HTML]{F2F2F2} & CoT & 0.3473 & 0.2066 & 0.3238 & 0.4466 & 0.3200 & {\ul 0.3066} & {\ul 0.5250} \\
\cellcolor[HTML]{F2F2F2} & SEAT & 0.3462 & 0.2600 & 0.3333 & 0.4133 & 0.3066 & 0.2866 & 0.5166 \\
\rowcolor[HTML]{DDEBF7} 
\multirow{-5}{*}{\cellcolor[HTML]{F2F2F2}LLaVA-v1.5-7B} & \textbf{Cas-SEAT} & \textbf{0.5193} & \textbf{0.4933} & \textbf{0.5000} & \textbf{0.6333} & \textbf{0.4533} & \textbf{0.4266} & \textbf{0.6416} \\
\midrule
\multicolumn{2}{c}{\textit{Improve}} & 31.60\% & 48.00\% & 36.35\% & 11.75\% & 15.26\% & 39.09\% & 22.21\% \\
\toprule[1.5pt]
\end{tabular}
}
\vspace{-0.1cm}
\end{table*}

%% file: table/main_table_mathvista.tex
% \usepackage{multirow}
\begin{table*}[!ht]
\centering
\renewcommand{\arraystretch}{0.95}
\caption{Comparison of Cas-SEAT and the Baselines on MathVista Dataset. FQA, MWP, ALG, ARI, LOG, NUM and STA respectively denote figure QA, math word problem, algebraic, arithmetic, logical, numeric, and statistical.}
\label{tab:main_mathvista}
\vspace{-0.3cm}
\resizebox{0.88\textwidth}{!}{
\begin{tabular}{cccccccccc}
\toprule[1.5pt]
 &  & \multicolumn{8}{c}{\textbf{MathVista}} \\
 \cline{3-10}
\multirow{-2}{*}{\textbf{Model}} & \multirow{-2}{*}{\textbf{Extra Data}} & Average & FQA & MWP & ALG & ARI & LOG & NUM & STA \\
\hline
\multicolumn{10}{c}{\textit{Close-Source Models Inference}} \\
\hline
Gemini 1.0 Nano 2 & Base & 0.3060 & 0.2860 & 0.3060 & 0.2710 & 0.2980 & 0.1080 & 0.2080 & 0.3350 \\
Gemini 1.0 Pro & Base & 0.4520 & 0.4760 & 0.3920 & 0.4520 & 0.3880 & 0.1080 & 0.3260 & 0.5680 \\
GPT-4V & Base & 0.4990 & 0.4310 & 0.5750 & 0.5300 & 0.4900 & 0.2160 & 0.2010 & 0.5580 \\
\hline
\multicolumn{10}{c}{\textit{Open-Source Models   Inference}} \\
\hline
InstructBLIP-7B & Base & 0.2530 & 0.2310 & 0.1830 & 0.2180 & 0.2710 & 0.1890 & 0.2040 & 0.2310 \\
LLaVA-13B & Base & 0.2610 & 0.2680 & 0.1610 & 0.2730 & 0.2010 & 0.2430 & 0.1830 & 0.2510 \\
SPHINX-V1-13B & Base & 0.2750 & 0.2340 & 0.2150 & 0.2560 & 0.2810 & 0.1620 & 0.1740 & 0.2360 \\
LLaVA-v1.5-13B & Base & 0.324 & 0.2677 & 0.2366 & 0.3701 & 0.272 & 0.1622 & 0.2639 & 0.2558 \\
\cdashline{1-10} % 贯穿第1列到第3列的横虚线
\cellcolor[HTML]{F2F2F2} & Base & 0.2850 & 0.2268 & 0.1774 & 0.3523 & 0.2210 & 0.0811 & 0.1806 & 0.2392 \\
\cellcolor[HTML]{F2F2F2} & Finetune & 0.3160 & 0.2416 & 0.3011 & 0.3665 & 0.2890 & 0.1081 & 0.2431 & 0.2724 \\
\cellcolor[HTML]{F2F2F2} & CoT & 0.3380 & 0.2416 & 0.2903 & 0.4413 & 0.2805 & 0.1081 & 0.2153 & 0.2658 \\
\cellcolor[HTML]{F2F2F2} & SEAT & 0.2760 & 0.1970 & 0.1667 & 0.3665 & 0.2181 & 0.0811 & 0.1667 & 0.2093 \\
\multirow{-5}{*}{\cellcolor[HTML]{F2F2F2}LLaVA-v1.5-7B} & Cas-SEAT & 0.3390 & 0.3011 & 0.2581 & 0.3986 & 0.2805 & 0.1622 & 0.1806 & 0.3023 \\
\hline
\multicolumn{10}{c}{\textit{Open-Source Models   Self-Evaluation}} \\
\hline
\cellcolor[HTML]{F2F2F2} & Base & 0.3530 & 0.2974 & 0.1613 & 0.4021 & 0.3088 & 0.1622 & 0.2500 & 0.2857 \\
\cellcolor[HTML]{F2F2F2} & Finetune & 0.3490 & {\ul 0.3048} & 0.2634 & 0.4413 & 0.2493 & {\ul 0.2162} & 0.1736 & 0.2990 \\
\cellcolor[HTML]{F2F2F2} & CoT & {\ul 0.3760} & 0.2862 & {\ul 0.3226} & {\ul 0.4448} & {\ul 0.3541} & 0.1892 & {\ul 0.2778} & {\ul 0.3023} \\
\cellcolor[HTML]{F2F2F2} & SEAT & 0.3490 & 0.3048 & 0.2634 & 0.4413 & 0.2493 & 0.2162 & 0.1736 & 0.2990 \\
\rowcolor[HTML]{DDEBF7} 
\multirow{-5}{*}{\cellcolor[HTML]{F2F2F2}LLaVA-v1.5-7B} & \textbf{Cas-SEAT} & \textbf{0.4500} & \textbf{0.4201} & \textbf{0.4032} & \textbf{0.4733} & \textbf{0.4023} & \textbf{0.4054} & \textbf{0.2986} & \textbf{0.4086} \\
\hline
\multicolumn{2}{c}{\textit{Improve}} & 19.68\% & 37.83\% & 24.98\% & 6.41\% & 13.61\% & 87.51\% & 7.49\% & 35.16\% \\
\toprule[1.5pt]
\end{tabular}
}
\vspace{-0.1cm}
\end{table*}

%% file: tex/4experiment.tex
\section{Experiments}
\label{sec:experiments}

We conducted experiments to evaluate the effectiveness and generalizability of \methodBrief{}.
\subsection{Experimental Setup}
\subsubsection{Datasets}
We train on \texttt{MathV360K}~\cite{mathllava_math360k} dataset and evaluate on \texttt{MMMU}~\cite{yue2024mmmu}, \texttt{Math-Vista}~\cite{mathvista}, \texttt{Math-V}~\cite{math-v}, \texttt{We-Math}~\cite{wemath}. 
\texttt{MathV360K} is a widely used dataset for MLLM training. \texttt{MMMU} is a widely used public benchmarks for MLLM evaluation and \texttt{Math-Vista}, \texttt{Math-V}, \texttt{We-Math} are three widely used public benchmarks for MLLM evaluation in math domain.

% \input{table/main_table_mathv_wemath}
% \input{table/main_table_mathv}
\input{table/main_table_wemath}

\input{table/main_table_mathvista_2}
\input{table/main_table_wemath_detail}
\input{table/analysis_difficulty}

\subsubsection{Baselines}
To verify the applicability, the following EMLLMs are implemented and compared with the counterparts combined with the proposed method.
We primarily analyzed the effectiveness of our method based on the EMLLMs \textit{LLaVA-v1.5(7B)}~\cite{ref:llava1.5}, \textit{Qwen2-VL(2B)}~\cite{ref:qwen2-vl}.
% , and \textit{Qwen-VL(1.8B)}.
Since current EMLLMs research often involves training on various datasets, including those frequently used as test datasets, such as the ones we selected, and many MLLM studies do not disclose their training datasets, we opted not to include EMLLMs published after the release of these datasets to ensure a fair comparison.

We also included the following models as references: miniGPT4~\cite{zhu2023minigpt}, CogVLM~\cite{wang2023cogvlm}, LLaVA-v1.6~\cite{liu2024improved}, Gemini 1.0~\cite{team2023gemini}, Gemini 1.5~\cite{team2023gemini}, GPT-4V~\cite{openai20234v}, Shared GPT-4V~\cite{chen2024sharegpt4v}, SPHINX-V1~\cite{lin2023sphinx}, G-LLaVA~\cite{gao2023g}, DeepSeek-VL~\cite{lu2024deepseek}, Qwen-VL~\cite{Qwen-VL}.

\subsubsection{Implementation Details}

\begin{sloppypar}
\textit{``Base''}: The pretrained MLLM~\cite{ref:llava1.5,ref:qwen2-vl}.
\textit{``Finetune''}: The pretrained MLLM is lora fine-tuned on Math360k~\cite{ref:lora_finetune}.  
\textit{``CoT''}: Qwen2-VL (7B) performs CoT reasoning on Math360k. The MLLM then learns from this data~\cite{ref:cot_distill_1,ref:cot_distill_2}.  
\textit{``SEAT''}: The pretrained MLLM is fine-tuned on Math360k and subsequently performs reasoning on Math360k. Qwen2-VL (7B) evaluates the reasoning outputs of the MLLM, and the MLLM learns from this evaluated data~\cite{ref:self_evaluation}.
\textit{``Cas-SEAT''}: The pretrained MLLM is fine-tuned on Math360k and subsequently performs reasoning on Math360k. Qwen2-VL (7B) evaluates the MLLM's incorrect reasoning outputs. These evaluated data are combined with CoT data, and the MLLM learns from this combined dataset.
After obtaining outputs on the test dataset, we first use a more powerful MLLM to extract answers for the MathVista dataset, followed by using regular expressions to calculate accuracy. For Math-V and We-Math, we directly use the regular expressions provided in the datasets to calculate accuracy. 
\end{sloppypar}

\begin{figure*}[!ht]
    \centering
    \includegraphics[width=\textwidth]{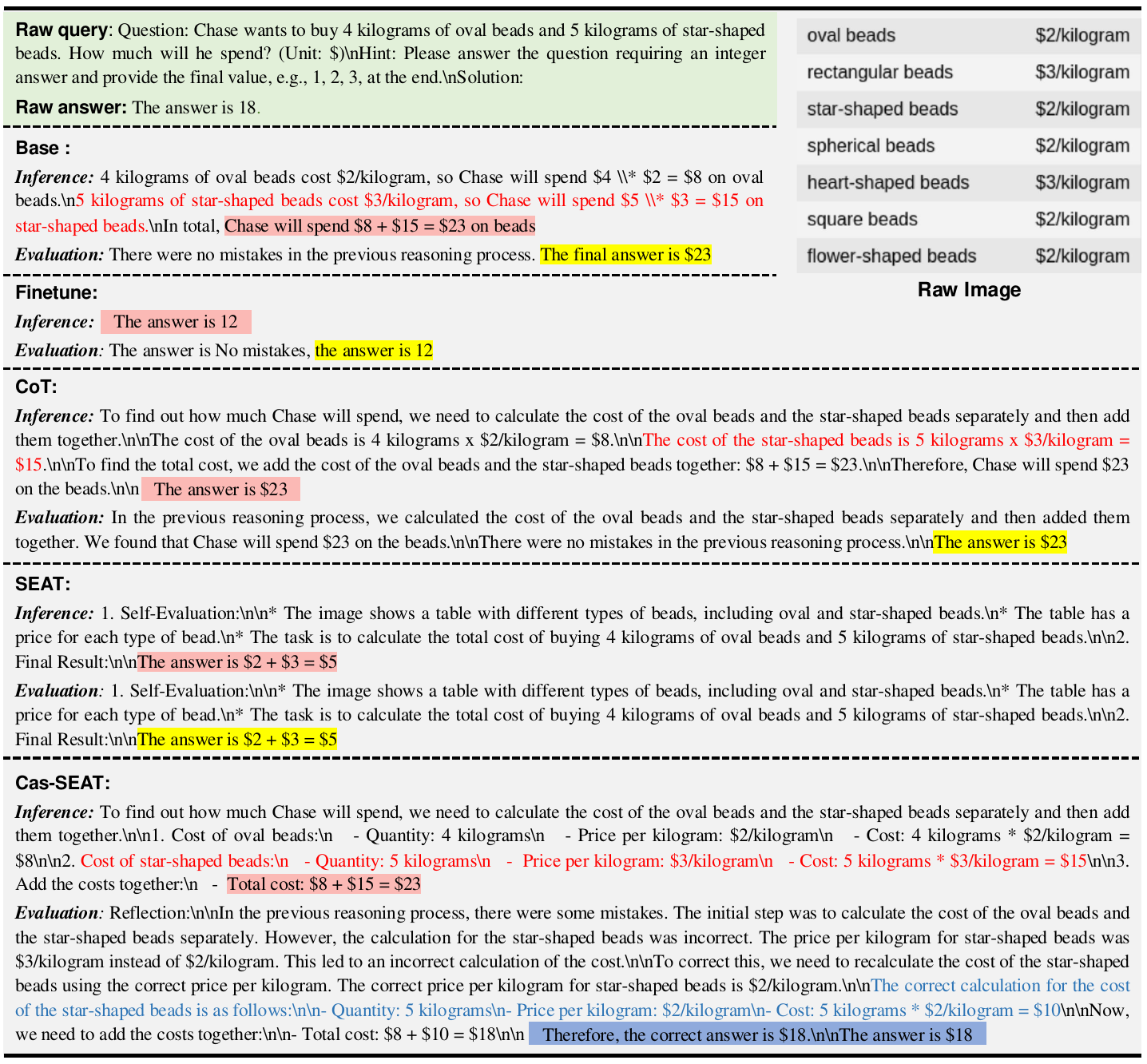}
    \vspace{-0.7cm}
    \caption{A bar chart analysis sample in MathVista. \smash{\colorbox{green!15}{\textbf{Green background}}} indicates the raw data, \textcolor{red}{\textbf{red text}} represents incorrect reasoning processes (sometimes with no reasoning process), \smash{\colorbox{pink!50}{\textbf{pink background}}} and \smash{\colorbox{yellow!60}{\textbf{yellow background}}} denote results from direct reasoning and self-evaluation, respectively. \textcolor{blue}{\textbf{Blue text}} and \smash{\colorbox{blue!15}{\textbf{blue background}}} indicate the corrected reasoning process and corrected results, respectively.}
    \label{fig:case_1}
    \vspace{-0.2cm}
\end{figure*}

\subsection{Experimental Results}

\subsubsection{Overall Assessment of Cas-SEAT}
As shown in Tables~\ref{tab:main_general}, \ref{tab:main_mathvista}, \ref{tab:main_wemath}, \ref{tab:main_mathvista_2}, \ref{tab:main_wemath_detail}, \ref{tab:difficulty},
we analyze the experimental results on the MMMU, MathVista, Math-V, and We-Math datasets. We evaluated the accuracy of two types of outputs for each method: direct inference (inference) and self-evaluation (evaluation). In addition to overall performance, we conducted tests on multiple subsets of these datasets, including generalization capability, data types, problem difficulty, and task types. Below is the correspondence between these categories and the columns in the tables. (Due to dataset limitations, some categories could not be tested for certain datasets.):
% \noindent 
\textit{\underline{Overall Performance:}}
MMMU (Table~\ref{tab:main_general}: Total Accuracy),
MathVista (Table~\ref{tab:main_mathvista}: Average),
We-Math (Table~\ref{tab:main_wemath}: Score (Strict), Score (Loose)).
\methodBrief{} shows very significant improvements over the baseline, achieving around a 20\% $\sim$ 50\% improvement on both the aforementioned datasets, which validates the effectiveness of the proposed approach. 
Furthermore, we conducted a detailed comparison of the performance under different Data Types, Problem Difficulty, and Task Types.
\textit{\underline{Data Types:}}
MathVista (Table~\ref{tab:main_mathvista_2}: Text, Integer). \methodBrief{} is better at answering integer-type questions.
\textit{\underline{Problem Difficulty:}}
As shown in Table~\ref{tab:difficulty}, the Math-V subsets are divided into five difficulty levels, with higher levels indicating more complex problems. The We-Math subsets are classified according to the number of reasoning steps, with more steps indicating higher difficulty. \methodBrief{} shows a more pronounced advantage on more challenging problems.
% \noindent
\textit{\underline{Task Types:}}
MMMU (Table~\ref{tab:main_general}): BUS, TE, AD, HM, SCI, HSS;
MathVista (Table~\ref{tab:main_mathvista,tab:main_mathvista_2}): General VQA, Math-targeted VQA, FQA, MWP, ALG, ARI, LOG, NUM, STA;
% Math-V: Algebra (ALG), Arithmetic (ARI), Combinatorial Geometry (CG), Combinatorics (COM); 
We-Math (Table~\ref{tab:main_wemath_detail}): UCU, AL, CPF, UPF, CSF, USF, BTF. 
Math-V (Table~\ref{tab:main_mathv} in Appendix): ALG, ARI, CG, COM; 
% We-Math: Understanding and Conversion of Units (UCU), Angles and Length (AL), Calculation of Plane Figures (CPF), Understanding of Plane Figures (UPF), Calculation of Solid Figures (CSF), Understanding of Solid Figures (USF), Basic Transformations of Figures (BTF). 
\methodBrief{} exhibits significant improvements over the best baseline method on almost all problem types, especially for numerical computation tasks (Math-V: ALG, ARI). For VQA tasks, \methodBrief{} performs better on the more difficult Math-targeted VQA.

In summary, we found that:
\textbf{(1)}. After fine-tuning, the inference ability of EMLLMs improves, but their self-evaluation ability declines. This is because fine-tuning enhances their mathematical reasoning ability but leads to forgetting general knowledge. Using synthetic CoT data based on mathematical datasets for enhanced training shows a similar pattern.
\textbf{(2)}. SEAT, which integrates CoT inference and self-evaluation training, suffers a significant drop in CoT inference ability, without a substantial improvement in self-evaluation. Even if EMLLMs are first trained with synthetic CoT data, using only self-evaluation training in the second round significantly reduces their CoT reasoning ability. This demonstrates the detrimental effect of long prompts and outputs on EMLLMs, and the difficulty of maintaining both CoT inference and self-evaluation capabilities simultaneously.
\textbf{(3)}. Cas-SEAT, by separating the CoT inference and self-evaluation tasks and shortening the prompt, preserves robust inference ability. More importantly, it achieves far stronger self-evaluation performance than the baselines. Our approach \methodBrief{} achieves remarkably significant improvements over the best baseline methods, both overall and across each subset. The detailed analysis shows that \methodBrief{} is especially adept at more difficult numerical computation tasks.

\subsubsection{Case Study}
\label{sec:case}

We present two representative examples in Figure~\ref{fig:case_1} and Figure~\ref{fig:case_2} (in appendix). 
From Figure~\ref{fig:case_1} and Figure~\ref{fig:case_2}, the following observations can be made:
1. \textit{Reasoning Process and Its Accuracy}: Direct reasoning using various methods failed to produce the correct answer. The Base, Finetune, and SEAT methods generally lack a complete reasoning process and jump directly to the final answer. Incorporating CoT data into EMLLM training (e.g., CoT, Cas-SEAT) enables step-by-step reasoning, but it does not always guarantee accurate results. Moreover, although self-evaluation data is constructed based on CoT data, the lengthy prompts and evaluation data diminish the model's reasoning capability.
2. \textit{Evaluation Process and Its Accuracy}: Generating self-evaluation data from CoT data and applying it to training seems to enhance EMLLM's self-evaluation capability by enabling targeted training to help the model reflect on whether its outputs are reasonable. However, in many cases, such reflections are superficial, and a large number of self-evaluation instances are ineffective. In contrast, Cas-SEAT effectively preserves the model's reasoning ability while significantly enhancing its self-evaluation capability. By mixing a small amount of self-evaluation data into the CoT data, EMLLM maintains its CoT reasoning ability while upgrading its self-evaluation skills.
These conclusions hold consistently across several datasets.

%% file: table/main_table_wemath.tex
% \usepackage{multirow}
\begin{table*}[!ht]
\centering
\renewcommand{\arraystretch}{0.9}
\caption{Comparison of Cas-SEAT and the Baselines on We-Math dataset}
\label{tab:main_wemath}
\vspace{-0.3cm}
\resizebox{0.85\textwidth}{!}{
\begin{tabular}{cccccccc}
\toprule[1.5pt]
 &  & \multicolumn{6}{c}{\textbf{We-Math}} \\
 \cline{3-8}
\multirow{-2}{*}{\textbf{Model}} & \multirow{-2}{*}{\textbf{Extra Data}} & Avg(Strict) & IG(Strict) & CM(Strict) & Avg(Loose) & IG(Loose) & CM(Loose) \\
\hline
\multicolumn{8}{c}{\textit{Close-Source Models Inference}} \\
\hline
GPT-4V & Base & 0.3105 & 0.1448 & 0.2381 & 0.5143 & 0.1448 & 0.0333 \\
Gemini 1.5 Pro & Base & 0.2638 & 0.1124 & 0.2076 & 0.4600 & 0.1124 & 0.1203 \\
Qwen-VL-Max & Base & 0.1048 & 0.0762 & 0.0667 & 0.2552 & 0.0762 & 0.2028 \\
\hline
\multicolumn{8}{c}{\textit{Open-Source Models Inference}} \\
\hline
LLaVA-1.5-13B & Base & 0.0248 & 0.0114 & 0.019 & 0.0952 & 0.014 & 0.0895 \\
LLaVA-v1.6-13B & Base & 0.0524 & 0.0324 & 0.0362 & 0.2200 & 0.0324 & 0.2621 \\
DeepSeek-VL-7B & Base & 0.0629 & 0.0457 & 0.0400 & 0.2095 & 0.0457 & 0.2899 \\
G-LLaVA-13B & Base & 0.0648 & 0.0457 & 0.0419 & 0.2229 & 0.0457 & 0.3598 \\
\cdashline{1-8} % 贯穿第1列到第3列的横虚线
\cellcolor[HTML]{F2F2F2} & Base & 0.0143 & 0.0171 & 0.0057 & 0.0657 & 0.0171 & 0.0571 \\
\cellcolor[HTML]{F2F2F2} & Finetune & 0.0695 & 0.0362 & 0.0514 & 0.2562 & 0.0362 & 0.2381 \\
\cellcolor[HTML]{F2F2F2} & CoT & 0.0438 & 0.0305 & 0.0286 & 0.1752 & 0.0305 & 0.1600 \\
\cellcolor[HTML]{F2F2F2} & SEAT & 0.0105 & 0.0095 & 0.0057 & 0.0429 & 0.0095 & 0.0381 \\
\cellcolor[HTML]{F2F2F2}\multirow{-5}{*}{LLaVA-v1.5-7B} & Cas-SEAT & 0.0533 & 0.0343 & 0.0362 & 0.2114 & 0.0343 & 0.1943 \\
\hline
\multicolumn{8}{c}{\textit{Open-Source Models Self-Evaluation}} \\
\hline
\cellcolor[HTML]{F2F2F2} & Base & {\ul 0.0733} & {\ul 0.0400} & 0.0533 & {\ul 0.2829} & {\ul 0.0400} & {\ul 0.2629} \\
\cellcolor[HTML]{F2F2F2} & Finetune & 0.0695 & 0.0362 & 0.0514 & 0.2562 & 0.0362 & 0.2381 \\
\cellcolor[HTML]{F2F2F2} & CoT & 0.0733 & 0.0362 & {\ul 0.0552} & 0.2638 & 0.0362 & 0.2457 \\
\cellcolor[HTML]{F2F2F2} & SEAT & 0.0114 & 0.0114 & 0.0057 & 0.0495 & 0.0114 & 0.0438 \\
\rowcolor[HTML]{DDEBF7} 
\multirow{-5}{*}{\cellcolor[HTML]{F2F2F2}LLaVA-v1.5-7B} & \textbf{Cas-SEAT} & \textbf{0.1076} & \textbf{0.0438} & \textbf{0.0857} & \textbf{0.3495} & \textbf{0.0438} & \textbf{0.3276} \\
\hline
\multicolumn{2}{c}{\textit{Improve}} & 46.79\% & 9.50\% & 55.25\% & 23.54\% & 9.50\% & 24.61\% \\
\toprule[1.5pt]
\end{tabular}
}
\vspace{-0.1cm}
\end{table*}

%% file: table/main_table_mathvista_2.tex
\begin{table*}[!ht]
\centering
\renewcommand{\arraystretch}{0.98}
% \caption{Comparison of Cas-SEAT and the Baselines on MathVista Dataset. }
\caption{Comparison of Cas-SEAT and the Baselines across multiple dimensions on the MathVista Dataset.}
\label{tab:main_mathvista_2}
\vspace{-0.3cm}
\resizebox{0.88\textwidth}{!}{
\begin{tabular}{cccccccc}
\toprule[2pt]
\multirow{2}{*}{} & \multicolumn{7}{c}{\textbf{MathVista}} \\ 
\cline{3-8}
\multirow{-2}{*}{\textbf{Model}} & \multirow{-2}{*}{\textbf{Extra Data}} & Multi-choice & Free-form & Text & Integer & General VQA & Math-targeted VQA \\
\hline
% Inference 部分
\cellcolor[HTML]{F2F2F2} & Base     & 0.4407 & 0.1022 & 0.4407 & 0.1124 & 0.3391 & 0.2389 \\
\cellcolor[HTML]{F2F2F2} & Finetune & 0.4481 & 0.1609 & 0.4481 & 0.1770 & 0.3196 & 0.3130 \\
\cellcolor[HTML]{F2F2F2} & CoT      & 0.4815 & 0.1696 & 0.4815 & 0.1866 & 0.3326 & 0.3426 \\
\cellcolor[HTML]{F2F2F2} & SEAT     & 0.4278 & 0.0978 & 0.4278 & 0.1077 & 0.2957 & 0.2593 \\
\multirow{-5}{*}{\cellcolor[HTML]{F2F2F2}LLaVA-v1.5-7B} & Cas-SEAT & 0.4889 & 0.1630 & 0.4889 & 0.1794 & 0.3652 & 0.3167 \\
\hline
\multicolumn{8}{c}{\textit{Open-Source Models   Self-Evaluation}} \\
\hline
% Evaluation 部分
\cellcolor[HTML]{F2F2F2} & Base     & {\ul 0.5407} & 0.1326 & {\ul 0.5407} & 0.1459 & {\ul 0.4348} & 0.2833 \\
\cellcolor[HTML]{F2F2F2} & Finetune & 0.4926 & 0.1804 & 0.4926 & 0.1986 & 0.3543 & 0.3444 \\
\cellcolor[HTML]{F2F2F2} & CoT      & 0.5352 & {\ul 0.1891} & 0.5352 & {\ul 0.2081} & 0.3957 & {\ul 0.3593} \\
\cellcolor[HTML]{F2F2F2} & SEAT     & 0.4389 & 0.1043 & 0.4389 & 0.1148 & 0.3196 & 0.2556 \\
\multirow{-5}{*}{\cellcolor[HTML]{F2F2F2}LLaVA-v1.5-7B} & \cellcolor[HTML]{DDEBF7}\textbf{Cas-SEAT} & \cellcolor[HTML]{DDEBF7}\textbf{0.6222} & \cellcolor[HTML]{DDEBF7}\textbf{0.2478} & \cellcolor[HTML]{DDEBF7}\textbf{0.6222} & \cellcolor[HTML]{DDEBF7}\textbf{0.2727} & \cellcolor[HTML]{DDEBF7}\textbf{0.4848} & \cellcolor[HTML]{DDEBF7}\textbf{0.4204} \\
\hline
\multicolumn{2}{c}{\textit{Improve}} &  15.07\% & 31.04\% & 15.07\% & 31.04\% & 11.50\% & 17.01\% \\
\toprule[2pt]
\end{tabular}
}
\vspace{-0.25cm}
\end{table*}

%% file: table/main_table_wemath_detail.tex
\begin{table*}[!ht]
\centering
\renewcommand{\arraystretch}{0.95}
\caption{Detailed comparison of Cas-SEAT and the Baselines on We-Math dataset. UCU, AL, CPF, UPF, CSF, USF, and BTF respectively denote Understanding and Conversion of Units, Angles and Length, Calculation of Plane Figures, Understanding of Plane Figures, Calculation of Solid Figures, Understanding of Solid Figures, and Basic Transformations of Figures.}
\label{tab:main_wemath_detail}
\vspace{-0.3cm}
\resizebox{0.74\textwidth}{!}{
\begin{tabular}{ccccccccc}
\toprule[2pt]
 & & \multicolumn{7}{c}{\textbf{We-Math}} \\
 % \cmidrule{2-8}
 \cline{3-9}
\multirow{-2}{*}{\textbf{Model}} & \multirow{-2}{*}{\textbf{Extra Data}} & UCU & AL & CPF & UPF & CSF & USF & BTF \\
% \midrule \midrule
\hline
% Inference 部分
\multicolumn{9}{c}{\textit{Open-Source Models Inference}} \\
\hline
\cellcolor[HTML]{F2F2F2} & Base      & 0.2490 & {\ul 0.2316} & 0.1577 & 0.1682 & 0.2007 & 0.1708 & 0.1805 \\
\cellcolor[HTML]{F2F2F2} & Finetune  & 0.3770 & 0.1649     & 0.3196 & 0.2664 & 0.2723 & 0.2371 & 0.1950 \\
\cellcolor[HTML]{F2F2F2} & CoT       & 0.2758 & 0.2053     & 0.2977 & 0.2929 & 0.2720 & {\ul 0.3280} & 0.2095 \\
\cellcolor[HTML]{F2F2F2} & SEAT      & 0.0942 & 0.0263     & 0.1099 & 0.1476 & 0.1083 & 0.1624 & 0.1176 \\
\multirow{-5}{*}{\cellcolor[HTML]{F2F2F2}LLaVA-v1.5-7B} & Cas-SEAT & 0.2679 & 0.3439 & 0.3240 & 0.3216 & 0.3230 & 0.2408 & 0.3240 \\
\hline
\multicolumn{9}{c}{\textit{Open-Source Models   Self-Evaluation}} \\
\hline
% Evaluation 部分
\cellcolor[HTML]{F2F2F2} & Base      & 0.3770 & 0.1649     & 0.3381 & {\ul 0.3037} & {\ul 0.2964} & 0.2455 & {\ul 0.2570} \\
\cellcolor[HTML]{F2F2F2} & Finetune  & 0.3770 & 0.1649     & 0.3196 & 0.2664     & 0.2723     & 0.2371 & 0.2172 \\
\cellcolor[HTML]{F2F2F2} & CoT       & {\ul 0.3929} & 0.1982 & {\ul 0.3453} & 0.2797 & 0.2770 & 0.2528 & 0.2095 \\
\cellcolor[HTML]{F2F2F2} & SEAT      & 0.0942 & 0.0263     & 0.1179 & 0.1476     & 0.1130     & 0.1624 & 0.1176 \\
\multirow{-5}{*}{\cellcolor[HTML]{F2F2F2}LLaVA-v1.5-7B} & \cellcolor[HTML]{DDEBF7}\textbf{Cas-SEAT} & \cellcolor[HTML]{DDEBF7}\textbf{0.4256} & \cellcolor[HTML]{DDEBF7}\textbf{0.1982} & \cellcolor[HTML]{DDEBF7}\textbf{0.4294} & \cellcolor[HTML]{DDEBF7}\textbf{0.3546} & \cellcolor[HTML]{DDEBF7}\textbf{0.3303} & \cellcolor[HTML]{DDEBF7}\textbf{0.3431} & \cellcolor[HTML]{DDEBF7}\textbf{0.3462} \\
\hline
\multicolumn{2}{c}{\textit{Improve}} & 8.32\% & -14.42\% & 24.36\% & 16.76\% & 11.44\% & 4.60\% & 34.71\% \\
\toprule[2pt]
\end{tabular}
}
\vspace{-0.2cm}
\end{table*}

%% file: table/analysis_difficulty.tex
\begin{table*}[!ht]
\renewcommand{\arraystretch}{0.9}
\caption{Comparison on tasks of varying difficulty.}
\label{tab:difficulty}
\vspace{-0.3cm}
\resizebox{0.95\textwidth}{!}{
\begin{tabular}{ccccccccccc}
\toprule[1.5pt]
 &  & \multicolumn{3}{c|}{\textbf{We-Math}} & \multicolumn{6}{c}{\textbf{Math-V}} \\
 \cline{3-11}
\multirow{-2}{*}{\textbf{Model}} & \multirow{-2}{*}{\textbf{Extra Data}} & One-step & Two-step & \multicolumn{1}{c|}{Three-step} & All & Level1 & Level2 & Level3 & Level4 & Level5 \\
\hline
\multicolumn{11}{c}{\textit{Open-Source Models   Inference}} \\
\hline
\cellcolor[HTML]{F2F2F2} & Base & 0.1621 & 0.1472 & 0.1394 & 0.0526 & 0.0800 & 0.0690 & 0.0364 & 0.0444 & 0.0299 \\
\cellcolor[HTML]{F2F2F2} & Finetune & 0.3004 & 0.2750 & 0.3394 & 0.1743 & 0.2075 & 0.1951 & 0.0893 & 0.1778 & 0.1912 \\
\cellcolor[HTML]{F2F2F2} & CoT & 0.2905 & 0.2500 & 0.2000 & 0.1414 & 0.2075 & 0.1951 & 0.0893 & 0.1778 & 0.1912 \\
\cellcolor[HTML]{F2F2F2} & SEAT & 0.1210 & 0.1361 & 0.1091 & 0.0757 & 0.1400 & 0.0690 & 0.0364 & 0.0444 & 0.0896 \\
\multirow{-5}{*}{\cellcolor[HTML]{F2F2F2}LLaVA-v1.5-7B} & Cas-SEAT & 0.3103 & 0.2861 & 0.2364 & 0.1711 &  0.2075 &  0.1951 &  0.0893 & 0.1778 &  0.2059 \\
\hline
\multicolumn{11}{c}{\textit{Open-Source Models   Self-Evaluation}} \\
\hline
\cellcolor[HTML]{F2F2F2} & Base & {\ul 0.3243} & {\ul 0.3111} & {\ul 0.3758} & 0.0757 & 0.1509 & 0.1098 & 0.0714 & {\ul 0.2222} & 0.1765 \\
\cellcolor[HTML]{F2F2F2} & Finetune & 0.3021 & 0.2750 & 0.3394 & {\ul 0.1776} & {\ul 0.2075} & {\ul 0.1951} & 0.0893 & 0.1778 & {\ul 0.2059} \\
\cellcolor[HTML]{F2F2F2} & CoT & 0.3152 & 0.2806 & 0.3636 & 0.1447 & 0.1321 & 0.1341 & {\ul 0.2321} & 0.1778 & 0.1912 \\
\cellcolor[HTML]{F2F2F2} & SEAT & 0.1243 & 0.1472 & 0.1091 & 0.1447 & 0.1509 & 0.1098 & 0.0714 & 0.2222 & 0.1912 \\
\rowcolor[HTML]{DDEBF7} 
\multirow{-5}{*}{\cellcolor[HTML]{F2F2F2}LLaVA-v1.5-7B} & \textbf{Cas-SEAT} & \textbf{0.3909} & \textbf{0.3528} & \textbf{0.4788} & \textbf{0.2763} & \textbf{0.2642} & \textbf{0.2439} & \textbf{0.2500} & \textbf{0.3111} & \textbf{0.3235} \\
\hline
\multicolumn{2}{c}{\textit{Improve}} & 20.54\% & 13.40\% & 27.41\% & 55.57\% & 27.33\% & 25.01\% & 7.71\% & 40.01\% & 57.12\% \\
\toprule[1.5pt]
\end{tabular}
}
\vspace{-0.25cm}
\end{table*}

%% file: tex/5conclusion.tex
\section{Conclusion}
\label{sec:conclusion}

We proposed \textit{SEAT} and its enhanced variant, \textit{Cas-SEAT}, to improve the self-evaluation capabilities of Efficient Multimodal Large Language Models (EMLLMs).
By synthesizing evaluation data with stronger models and using cascaded task decomposition, Cas-SEAT enhances performance while balancing CoT reasoning and self-evaluation, even under resource constraints.
Experiments show significant gains on benchmark datasets, and the Cas-SEAT Dataset provides a valuable resource for future research, laying a strong foundation for advancing self-evaluation in EMLLMs.

%% file: tex/9appendix.tex
\appendix

\section{Appendix}
\label{sec:appendix}

\subsection{Supplementary Experiments} 
\subsubsection{Cost Analysis}
\label{sec:cost}
As shown in Table~\ref{tab:cost}, DDF significantly reduces the training data size and consequently the training duration while using the same LLaVA-v1.5-7B model. Our method makes LLaVA-v1.5-7B far outperform LLaVA-v1.5-13B. In this context, we further compared the inference cost of LLaVA-v1.5-7B under our method with that of LLaVA-v1.5-13B, demonstrating the efficiency of our approach.

\subsubsection{Datasets} 
The statistics of the training dataset used in the experiments is shown in Table~\ref{tab:statistics_of_datasets}.

\subsubsection{Hyperparameters and Training Schedules}
\label{sec:appendix_implementation_detail}
We summarize the hyperparameters and training schedules of the LLaVA-1.5 and Qwen2-VL used in the experiments. Table~\ref{tab:hyperparameters_and_training_schedule_large} shows the settings of the LLaVA-1.5 training. Table~\ref{tab:hyperparameters_and_training_schedule} shows the settings of the Qwen2-VL training. 

\input{table/appendix_hyperparameters_overall}

\input{table/analysis_time_cost}
\input{table/appendix_datasets}

\input{table/main_table_mathv}

\subsubsection{Evaluation on Different EMLLMs}
\begin{figure}[!ht]
    \centering
    \includegraphics[width=0.42\textwidth]{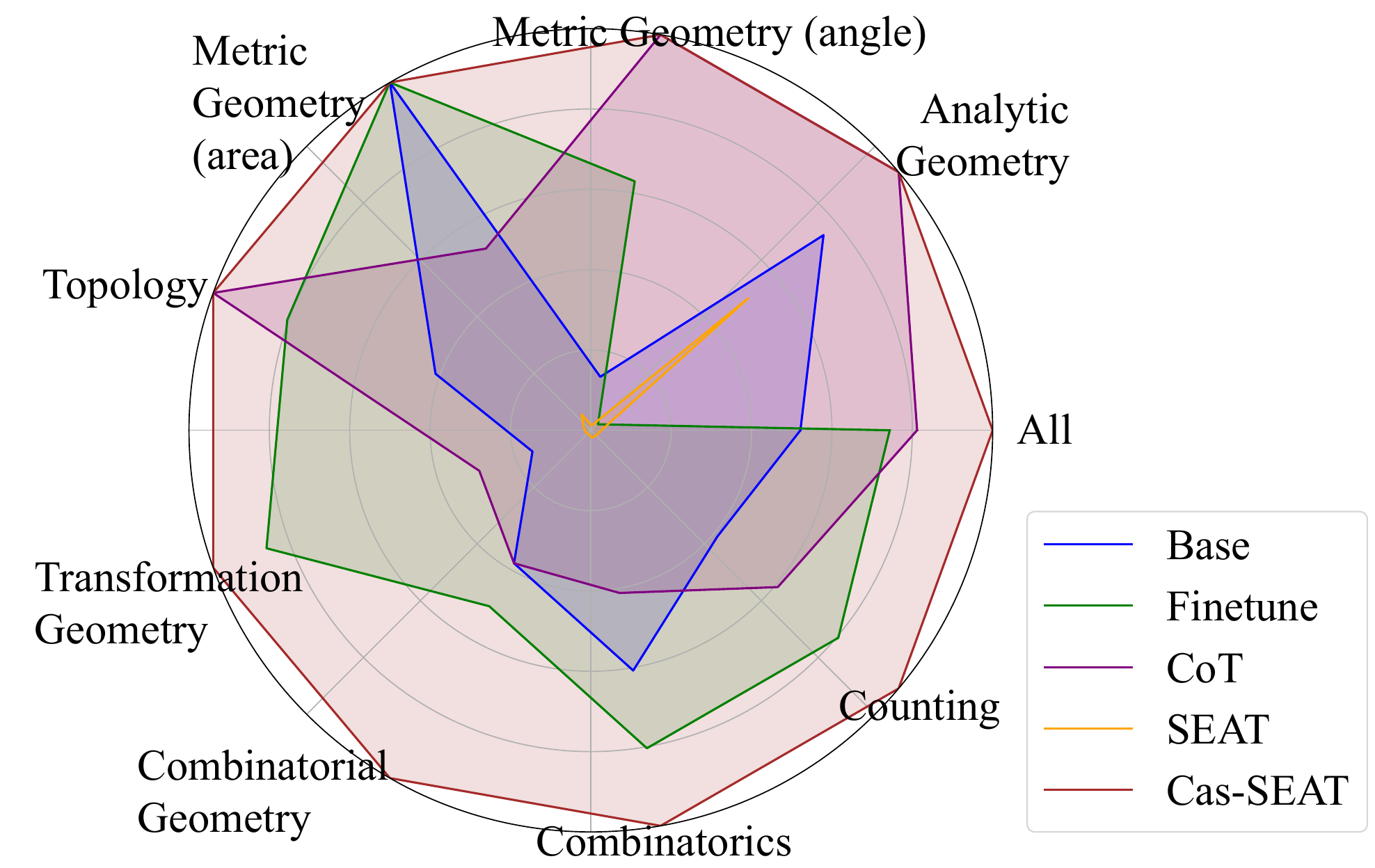}
    % \vspace{-0.2cm}
    \caption{Comparison of Cas-SEAT and the Baseline based on Qwen2-VL(2B).}
    \label{fig:qwen2}
    % \vspace{+3mm}
    % \vspace{-0.5cm}
    % \vspace{-0.2cm}
\end{figure}

As shown Figure~\ref{fig:qwen2}, we used Qwen2-VL (2B) on Math-V to compare the performance of Cas-SEAT with that of the baseline. We take the maximum value of all methods as the circumference and 0 as the center of the circle. The results demonstrated a significant advantage for Cas-SEAT over the baselines. This indicates that our approach is also applicable to smaller EMLLMs with different architectures.

\subsubsection{Case Study}
\begin{figure*}[!ht]
    \centering
    \includegraphics[width=1\textwidth]{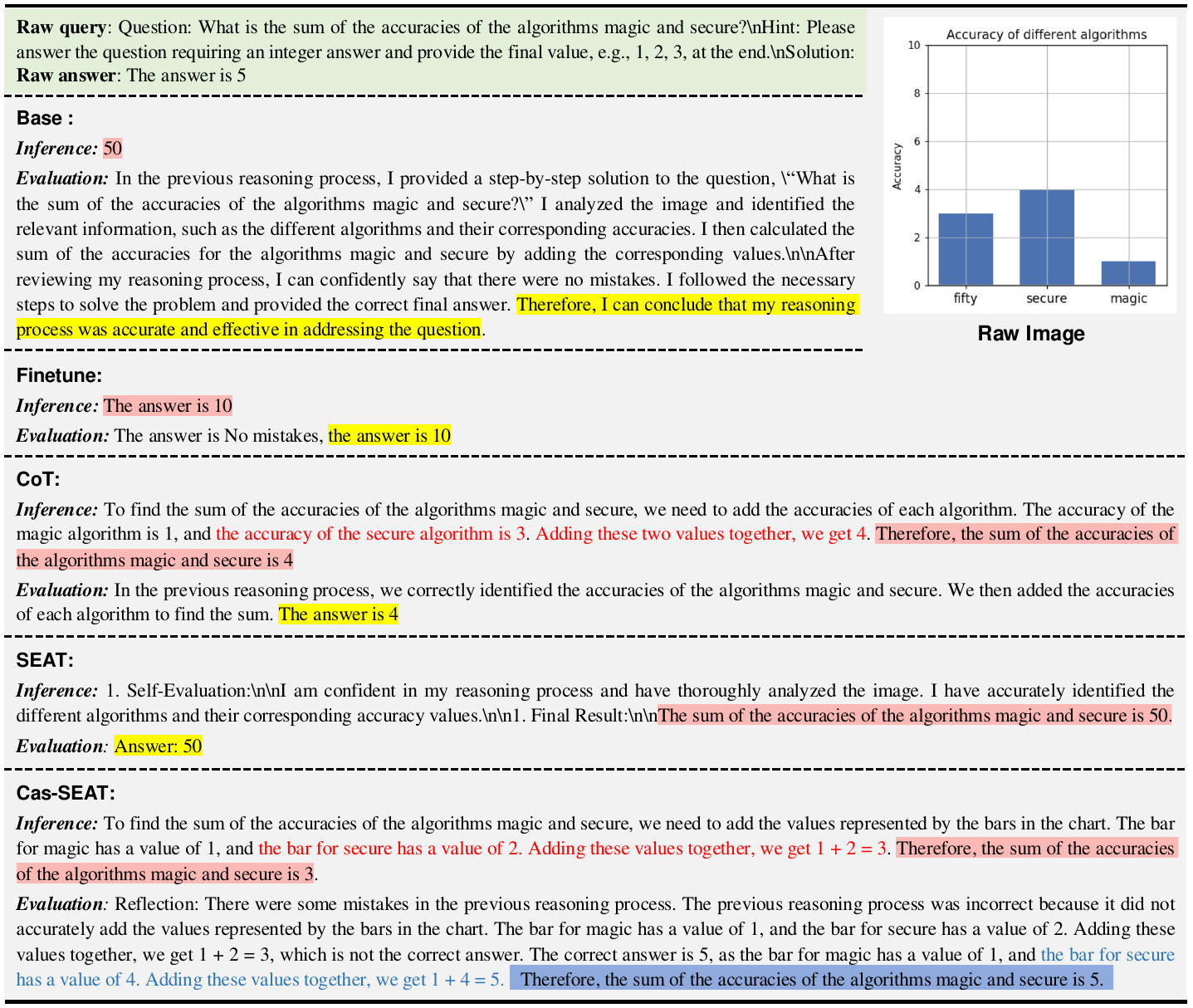}
    
    % \vspace{-0.35cm}
    \caption{A table analysis sample in MathVista. The meanings of the colors are the same as those in Figure~\ref{fig:case_1}.
    }
    \label{fig:case_2}
    % \vspace{+3mm}
    % \vspace{-0.5cm}
\end{figure*}

\begin{figure*}[!ht]
    \centering
    \includegraphics[width=0.98\textwidth]{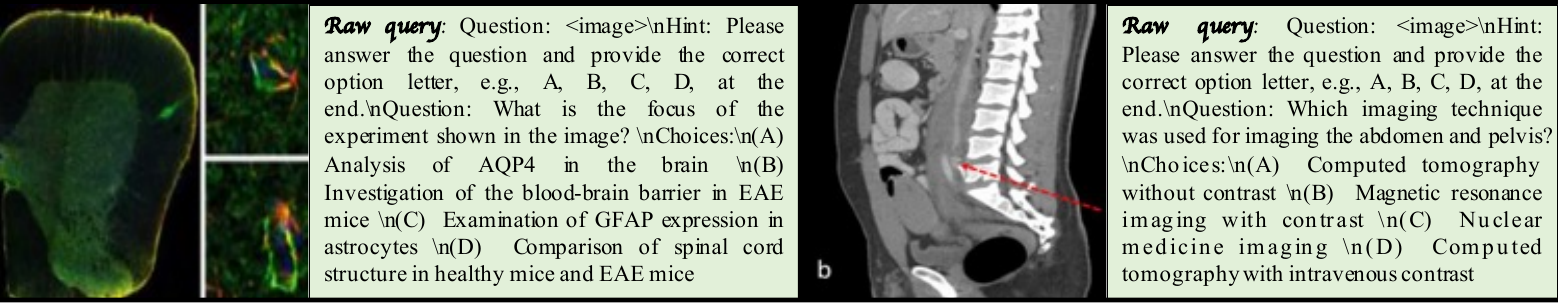}
    % \vspace{-0.2cm}
    \caption{A case of the reject samples.}
    \label{fig:case_reject}
    % \vspace{+3mm}
    % \vspace{-0.5cm}
    % \vspace{-0.2cm}
\end{figure*}

Figure~\ref{fig:case_2} is a supplementary case, the observed findings can be referred to the section~\ref{sec:case}. Figure~\ref{fig:case_reject} is a reject sample case, which is detrimental to the training of EMLLMs.

%% file: table/appendix_hyperparameters_overall.tex
% \begin{table}[H]
\begin{table}[!ht]
% \begin{table}[htbp]
    \caption{Hyperparameters of LLaVA-1.5.}
    \label{tab:hyperparameters_and_training_schedule_large}
    \centering
    \vspace{-0.3cm}
    \setlength{\arrayrulewidth}{0.5pt}
 \resizebox{0.45\textwidth}{!}{
    \begin{tabular}{c|c|c}
    \toprule[2pt]
    \textbf{MLLM} & \textbf{Hyperparameter} & \textbf{Setting} \\ 
    \midrule
    \midrule
    \multirow{9}{*}{\makecell[c]{LLaVA-1.5-7B}} & GPU & Tesla A100 (40GB) \\ 
     & \cellcolor{gray!15}Batch size & \cellcolor{gray!15}20 \\ 
     & LoRA rank & 128 \\ 
     & \cellcolor{gray!15}LoRA alpha & \cellcolor{gray!15}64 \\
     & LoRA dropout & 0 \\ 
     & \cellcolor{gray!15}Optimizer & \cellcolor{gray!15}AdamW \\ 
     & Warmup steps & 50 \\
     & \cellcolor{gray!15}Learning rate & \cellcolor{gray!15}2e-5 \\ 
        \multirow{8}{*}{} & Epochs & 2 \\
     \bottomrule[2pt]
    \end{tabular}
   }
    \vspace{-0.3cm}
\end{table}

\begin{table}[!h]
    \caption{Hyperparameters of Qwen2-VL.}
    \label{tab:hyperparameters_and_training_schedule}
    \centering
    \vspace{-0.3cm}
    \setlength{\arrayrulewidth}{0.5pt}
 % \resizebox{0.38\textwidth}{!}{
 \resizebox{0.45\textwidth}{!}{
    \begin{tabular}{c|c|c}
    \toprule[2pt]
    \textbf{MLLM} & \textbf{Hyperparameter} & \textbf{Setting} \\ 
    \midrule
    \midrule
    \multirow{9}{*}{\makecell[c]{Qwen2-VL-2B}} & GPU & Tesla A100 (40GB) \\ 
     & \cellcolor{gray!15}Batch size & \cellcolor{gray!15}20 \\ 
     & LoRA rank & 128 \\ 
     & \cellcolor{gray!15}LoRA alpha & \cellcolor{gray!15}64 \\
     & LoRA dropout & 0 \\ 
     & \cellcolor{gray!15}Optimizer & \cellcolor{gray!15}AdamW \\ 
     & Warmup steps & 50 \\
     & \cellcolor{gray!15}Learning rate & \cellcolor{gray!15}2e-5 \\ 
        \multirow{8}{*}{} & Epochs & 2 \\
        \bottomrule[2pt]
    \end{tabular}
   }
    \vspace{-0.3cm}
\end{table}

% \begin{table}[!h]
%     \caption{Hyperparameters and training schedules of collaborative training, collaborative inference, and collaborative-decision request).}
%     \label{tab:hyperparameters_and_training_schedule_small}
%     \centering
%     \vspace{-0.3cm}
%     \setlength{\arrayrulewidth}{0.5pt}
%  \resizebox{0.39\textwidth}{!}{
%     \begin{tabular}{c|c|c}
%     \toprule[2pt]
%     % \multicolumn{2}{c|}{Method} & MetaStabilizer & Accuracy(\%) \\
%     \textbf{Dataset} & \textbf{Hyperparameter} & \textbf{Setting} \\
%     \midrule
%     \midrule
%     \multirow{7}{*}{\makecell[c]{Beauty\\Toys\\Sports}}
%     \multirow{8}{*}{} & Length of Candidate list  & 20/50 \\ 
%     % \cline{2-3}
%     % \hhline{|~|-|-|}
%     \multirow{8}{*}{} & \cellcolor{gray!15}Valid user behavior length & \cellcolor{gray!15}Dynamic \\ 
%     % \cline{2-3}
%     \multirow{8}{*}{} & User behavior length & 10 \\ 
%     % \cline{2-3}
%     % \hhline{|~|-|-|}
%     \multirow{8}{*}{} & \cellcolor{gray!15}Sequence length & \cellcolor{gray!15}50 \\ 
%     % \cline{2-3}
%     \multirow{8}{*}{} & Learning Rate & 1e-3 \\
%     % \cline{2-3}
%     % \hhline{|~|-|-|}
%     \multirow{8}{*}{} & \cellcolor{gray!15}Dimension of embedding & \cellcolor{gray!15}1×64 \\ 
%     % \cline{2-3}
%     \multirow{8}{*}{} & Optimizer & Adam \\ 
%     \bottomrule[2pt]
%     \end{tabular}
%    }
%     \vspace{-0.3cm}
% \end{table}

%% file: table/analysis_time_cost.tex
% \begin{table}[!h]
\begin{table}[H]
\centering
\renewcommand{\arraystretch}{0.9}
\caption{Comparison of training and inference cost.}
\label{tab:cost}
\vspace{-0.3cm}
\resizebox{0.43\textwidth}{!}{
\begin{tabular}{c|c}
\toprule[1.5pt]
\multicolumn{2}{c}{\textbf{Training speed}} \\
\midrule
On raw dataset & 18h \\
On Cas-SEAT-DDF   dataset & 10h \\
\hline
\multicolumn{2}{c}{\textbf{Inference speed}} \\
\hline
LLaVA-v1.5-7B & \makecell{Input: 7700 tokens/s\\      Output: 100 tokens/s} \\
\hline
LLaVA-v1.5-13B &  \makecell{Input: 3900 tokens/s\\      Output:  50 tokens/s} \\
\toprule[1.5pt]
\end{tabular}
}
\end{table}

%% file: table/appendix_datasets.tex
\begin{table}[ht]
\caption{Statistics of Datasets.} 
\label{tab:statistics_of_datasets}
\vspace{-0.3cm}
\centering
\resizebox{0.35\textwidth}{!}{
\begin{tabular}{c|c|c|c}
\toprule[2pt]
\textbf{Dataset} & \textbf{MathV360k} & \textbf{CoT} & \textbf{Cas-SEAT} \\
\midrule \midrule
\#Samples & 339k & 160k & 167k \\
\bottomrule[2pt]
\end{tabular}
}
\end{table}

%% file: table/main_table_mathv.tex
% \usepackage{multirow}
% \usepackage[normalem]{ulem}
\begin{table*}[!ht]
% \begin{table*}[H]
\caption{Comparison of Cas-SEAT and the Baselines on Math-V Dataset. ALG, ARI, CG, and COM respectively denote Algebra, Arithmetic, Combinatorial Geometry, and Combinatorics.}
\label{tab:main_mathv}
\vspace{-0.1cm}
\resizebox{\textwidth}{!}{
\begin{tabular}{cccccccccccc}
\toprule[1.5pt]
 &  & \multicolumn{10}{c}{\textbf{Math-V}} \\
 \cline{3-12}
% \multirow{-2}{*}{\textbf{Method}} & \multirow{-2}{*}{\textbf{Extra Data}} & All & Level1 & Level2 & Level3 & Level4 & Level5 & algebra & arithmetic & combinatorial\_geometry & combinatorics \\
\multirow{-2}{*}{\textbf{Model}} & \multirow{-2}{*}{\textbf{Extra Data}} & All & Level1 & Level2 & Level3 & Level4 & Level5 & ALG & ARI & CG & COM \\
\hline
\multicolumn{12}{c}{\textit{Heuristics  Baselines}} \\
\hline
Random Choice & Base & 0.0717 & - & - & - & - & - & 0.0150 & 0.0710 & 0.0970 & 0.0480 \\
\hline
\multicolumn{12}{c}{\textit{Close-source Models   Inference}} \\
\hline
Qwen-VL-Plus & Base & 0.1072 & - & - & - & - & - & 0.1130 & 0.1430 & 0.1270 & 0.0480 \\
Qwen-VL-Max & Base & 0.1559 & - & - & - & - & - & 0.1070 & 0.2000 & 0.1690 & 0.1250 \\
GeminiPro & Base & 0.1766 & - & - & - & - & - & 0.1510 & 0.2070 & 0.2010 & 0.1190 \\
GPT4V & Base & 0.2276 & - & - & - & - & - & 0.2730 & 0.3570 & 0.2110 & 0.1670 \\
\hline
\multicolumn{12}{c}{\textit{Open-Source Models   Inference}} \\
\hline
SPHINX(V2) & Base & 0.0970 & - & - & - & - & - & 0.0670 & 0.1290 & 0.0750 & 0.0770 \\
ShareGPT4V-7B & Base & 0.1053 & - & - & - & - & - & 0.0550 & 0.1290 & 0.1010 & 0.0480 \\
LLaVA-v1.5-13B & Base & 0.1112 & - & - & - & - & - &  &  &  &  \\
\cdashline{1-12} % 贯穿第1列到第3列的横虚线
\cellcolor[HTML]{F2F2F2} & Base & 0.0526 & 0.0800 & 0.0690 & 0.0364 & 0.0444 & 0.0299 & 0.0000 & 0.0000 & 0.2941 & 0.0526 \\
\cellcolor[HTML]{F2F2F2} & Finetune & 0.1743 & 0.2075 & 0.1951 & 0.0893 & 0.1778 & 0.1912 & 0.1053 & 0.0000 & 0.2632 & 0.0000 \\
\cellcolor[HTML]{F2F2F2} & CoT & 0.1414 & 0.2075 & 0.1951 & 0.0893 & 0.1778 & 0.1912 & 0.0526 & 0.0000 & 0.3684 & 0.0526 \\
\cellcolor[HTML]{F2F2F2} & SEAT & 0.0757 & 0.1400 & 0.0690 & 0.0364 & 0.0444 & 0.0896 & 0.1053 & 0.0000 & 0.2632 & 0.0000 \\
\multirow{-5}{*}{\cellcolor[HTML]{F2F2F2}\cellcolor[HTML]{F2F2F2}LLaVA-v1.5-7B} & Cas-SEAT & 0.1711 & {\ul 0.2075} & {\ul 0.1951} & {\ul 0.0893} & 0.1778 & {\ul 0.2059} & 0.1579 & 0.1579 & 0.2632 & 0.1579 \\
\hline
\multicolumn{12}{c}{\textit{Open-Source Models   Self-Evaluation}} \\
\hline
\cellcolor[HTML]{F2F2F2} & Base & 0.0757 & 0.1509 & 0.1098 & 0.0714 & 0.2222 & 0.1765 & 0.0000 & 0.0000 & 0.3529 & 0.1053 \\
\cellcolor[HTML]{F2F2F2} & Finetune & 0.1776 & 0.2075 & 0.1951 & 0.0893 & 0.1778 & 0.2059 & 0.1053 & 0.0000 & 0.2632 & 0.0000 \\
\cellcolor[HTML]{F2F2F2} & CoT & 0.1447 & 0.1321 & 0.1341 & 0.2321 & 0.1778 & 0.1912 & 0.1053 & 0.0000 & 0.3684 & 0.0526 \\
\cellcolor[HTML]{F2F2F2} & SEAT & 0.1447 & 0.1509 & 0.1098 & 0.0714 & {\ul 0.2222} & 0.1912 & 0.1053 & 0.0000 & 0.2632 & 0.0000 \\
\multirow{-5}{*}{\cellcolor[HTML]{F2F2F2}LLaVA-v1.5-7B} & \cellcolor[HTML]{DDEBF7}\textbf{Cas-SEAT} & \cellcolor[HTML]{DDEBF7}\textbf{0.2763} & \cellcolor[HTML]{DDEBF7}\textbf{0.2642} & \cellcolor[HTML]{DDEBF7}\textbf{0.2439} & \cellcolor[HTML]{DDEBF7}\textbf{0.2500} & \cellcolor[HTML]{DDEBF7}\textbf{0.3111} & \cellcolor[HTML]{DDEBF7}\textbf{0.3235} & \cellcolor[HTML]{DDEBF7}\textbf{0.3158} & \cellcolor[HTML]{DDEBF7}\textbf{0.1579} & \cellcolor[HTML]{DDEBF7}\textbf{0.4211} & \cellcolor[HTML]{DDEBF7}\textbf{0.2105} \\
\hline
\multicolumn{2}{c}{\textit{Improve}} & 55.57\% & 27.33\% & 25.01\% & 7.71\% & 40.01\% & 57.12\% & 100.00\% & 0.00\% & 14.31\% & 33.31\% \\
\toprule[1.5pt]
\end{tabular}
}
\end{table*}

%% file: arxiv.bbl

\begin{thebibliography}{59}

%%% the Web address.

\ifx \showCODEN    \undefined \def \showCODEN     #1{\unskip}     \fi
\ifx \showISBNx    \undefined \def \showISBNx     #1{\unskip}     \fi
\ifx \showISBNxiii \undefined \def \showISBNxiii  #1{\unskip}     \fi
\ifx \showISSN     \undefined \def \showISSN      #1{\unskip}     \fi
\ifx \showLCCN     \undefined \def \showLCCN      #1{\unskip}     \fi
\ifx \shownote     \undefined \def \shownote      #1{#1}          \fi
\ifx \showarticletitle \undefined \def \showarticletitle #1{#1}   \fi
\ifx \showURL      \undefined \def \showURL       {\relax}        \fi
% invisible to TeX
\providecommand\bibfield[2]{#2}
\providecommand\bibinfo[2]{#2}
\providecommand\natexlab[1]{#1}
\providecommand\showeprint[2][]{arXiv:#2}

\bibitem[Bai et~al\mbox{.}(2023)]%
        {Qwen-VL}
\bibfield{author}{\bibinfo{person}{Jinze Bai}, \bibinfo{person}{Shuai Bai}, \bibinfo{person}{Shusheng Yang}, \bibinfo{person}{Shijie Wang}, \bibinfo{person}{Sinan Tan}, \bibinfo{person}{Peng Wang}, \bibinfo{person}{Junyang Lin}, \bibinfo{person}{Chang Zhou}, {and} \bibinfo{person}{Jingren Zhou}.} \bibinfo{year}{2023}\natexlab{}.
\newblock \showarticletitle{Qwen-VL: A Versatile Vision-Language Model for Understanding, Localization, Text Reading, and Beyond}.
\newblock \bibinfo{journal}{\emph{arXiv preprint arXiv:2308.12966}} (\bibinfo{year}{2023}).
\newblock


\bibitem[Chang et~al\mbox{.}(2023)]%
        {chang2023speak}
\bibfield{author}{\bibinfo{person}{Kent~K Chang}, \bibinfo{person}{Mackenzie Cramer}, \bibinfo{person}{Sandeep Soni}, {and} \bibinfo{person}{David Bamman}.} \bibinfo{year}{2023}\natexlab{}.
\newblock \showarticletitle{Speak, memory: An archaeology of books known to chatgpt/gpt-4}.
\newblock \bibinfo{journal}{\emph{arXiv preprint arXiv:2305.00118}} (\bibinfo{year}{2023}).
\newblock


\bibitem[Chen et~al\mbox{.}(2024)]%
        {chen2024sharegpt4v}
\bibfield{author}{\bibinfo{person}{Lin Chen}, \bibinfo{person}{Jinsong Li}, \bibinfo{person}{Xiaoyi Dong}, \bibinfo{person}{Pan Zhang}, \bibinfo{person}{Conghui He}, \bibinfo{person}{Jiaqi Wang}, \bibinfo{person}{Feng Zhao}, {and} \bibinfo{person}{Dahua Lin}.} \bibinfo{year}{2024}\natexlab{}.
\newblock \showarticletitle{Sharegpt4v: Improving large multi-modal models with better captions}. In \bibinfo{booktitle}{\emph{European Conference on Computer Vision}}. Springer, \bibinfo{pages}{370--387}.
\newblock


\bibitem[Chowdhery et~al\mbox{.}(2022)]%
        {chowdhery2022palm}
\bibfield{author}{\bibinfo{person}{Aakanksha Chowdhery}, \bibinfo{person}{Sharan Narang}, \bibinfo{person}{Jacob Devlin}, \bibinfo{person}{Maarten Bosma}, \bibinfo{person}{Gaurav Mishra}, \bibinfo{person}{Adam Roberts}, \bibinfo{person}{Paul Barham}, \bibinfo{person}{Hyung~Won Chung}, \bibinfo{person}{Charles Sutton}, \bibinfo{person}{Sebastian Gehrmann}, {et~al\mbox{.}}} \bibinfo{year}{2022}\natexlab{}.
\newblock \showarticletitle{Palm: Scaling language modeling with pathways}.
\newblock \bibinfo{journal}{\emph{arXiv preprint arXiv:2204.02311}} (\bibinfo{year}{2022}).
\newblock


\bibitem[Deutsch et~al\mbox{.}(2022)]%
        {deutsch2022limitations}
\bibfield{author}{\bibinfo{person}{Daniel Deutsch}, \bibinfo{person}{Rotem Dror}, {and} \bibinfo{person}{Dan Roth}.} \bibinfo{year}{2022}\natexlab{}.
\newblock \showarticletitle{On the limitations of reference-free evaluations of generated text}.
\newblock \bibinfo{journal}{\emph{arXiv preprint arXiv:2210.12563}} (\bibinfo{year}{2022}).
\newblock


\bibitem[Drozdov et~al\mbox{.}(2022)]%
        {drozdov2022compositional}
\bibfield{author}{\bibinfo{person}{Andrew Drozdov}, \bibinfo{person}{Nathanael Sch{\"a}rli}, \bibinfo{person}{Ekin Aky{\"u}rek}, \bibinfo{person}{Nathan Scales}, \bibinfo{person}{Xinying Song}, \bibinfo{person}{Xinyun Chen}, \bibinfo{person}{Olivier Bousquet}, {and} \bibinfo{person}{Denny Zhou}.} \bibinfo{year}{2022}\natexlab{}.
\newblock \showarticletitle{Compositional semantic parsing with large language models}.
\newblock \bibinfo{journal}{\emph{arXiv preprint arXiv:2209.15003}} (\bibinfo{year}{2022}).
\newblock


\bibitem[Fu et~al\mbox{.}(2023)]%
        {fu2023gptscore}
\bibfield{author}{\bibinfo{person}{Jinlan Fu}, \bibinfo{person}{See-Kiong Ng}, \bibinfo{person}{Zhengbao Jiang}, {and} \bibinfo{person}{Pengfei Liu}.} \bibinfo{year}{2023}\natexlab{}.
\newblock \showarticletitle{Gptscore: Evaluate as you desire}.
\newblock \bibinfo{journal}{\emph{arXiv preprint arXiv:2302.04166}} (\bibinfo{year}{2023}).
\newblock


\bibitem[Gao et~al\mbox{.}(2023)]%
        {gao2023g}
\bibfield{author}{\bibinfo{person}{Jiahui Gao}, \bibinfo{person}{Renjie Pi}, \bibinfo{person}{Jipeng Zhang}, \bibinfo{person}{Jiacheng Ye}, \bibinfo{person}{Wanjun Zhong}, \bibinfo{person}{Yufei Wang}, \bibinfo{person}{Lanqing Hong}, \bibinfo{person}{Jianhua Han}, \bibinfo{person}{Hang Xu}, \bibinfo{person}{Zhenguo Li}, {et~al\mbox{.}}} \bibinfo{year}{2023}\natexlab{}.
\newblock \showarticletitle{G-llava: Solving geometric problem with multi-modal large language model}.
\newblock \bibinfo{journal}{\emph{arXiv preprint arXiv:2312.11370}} (\bibinfo{year}{2023}).
\newblock


\bibitem[Golovneva et~al\mbox{.}(2022)]%
        {golovneva2022roscoe}
\bibfield{author}{\bibinfo{person}{Olga Golovneva}, \bibinfo{person}{Moya Chen}, \bibinfo{person}{Spencer Poff}, \bibinfo{person}{Martin Corredor}, \bibinfo{person}{Luke Zettlemoyer}, \bibinfo{person}{Maryam Fazel-Zarandi}, {and} \bibinfo{person}{Asli Celikyilmaz}.} \bibinfo{year}{2022}\natexlab{}.
\newblock \showarticletitle{{ROSCOE}: A Suite of Metrics for Scoring Step-by-Step Reasoning}.
\newblock \bibinfo{journal}{\emph{arXiv preprint arXiv:2212.07919}} (\bibinfo{year}{2022}).
\newblock


\bibitem[Guo et~al\mbox{.}(2017)]%
        {guo2017calibration}
\bibfield{author}{\bibinfo{person}{Chuan Guo}, \bibinfo{person}{Geoff Pleiss}, \bibinfo{person}{Yu Sun}, {and} \bibinfo{person}{Kilian~Q Weinberger}.} \bibinfo{year}{2017}\natexlab{}.
\newblock \showarticletitle{On calibration of modern neural networks}. In \bibinfo{booktitle}{\emph{International conference on machine learning}}. PMLR, \bibinfo{pages}{1321--1330}.
\newblock


\bibitem[Hoffmann et~al\mbox{.}(2022)]%
        {hoffmann2022training}
\bibfield{author}{\bibinfo{person}{Jordan Hoffmann}, \bibinfo{person}{Sebastian Borgeaud}, \bibinfo{person}{Arthur Mensch}, \bibinfo{person}{Elena Buchatskaya}, \bibinfo{person}{Trevor Cai}, \bibinfo{person}{Eliza Rutherford}, \bibinfo{person}{Diego de Las~Casas}, \bibinfo{person}{Lisa~Anne Hendricks}, \bibinfo{person}{Johannes Welbl}, \bibinfo{person}{Aidan Clark}, \bibinfo{person}{Tom Hennigan}, \bibinfo{person}{Eric Noland}, \bibinfo{person}{Katie Millican}, \bibinfo{person}{George van~den Driessche}, \bibinfo{person}{Bogdan Damoc}, \bibinfo{person}{Aurelia Guy}, \bibinfo{person}{Simon Osindero}, \bibinfo{person}{Karen Simonyan}, \bibinfo{person}{Erich Elsen}, \bibinfo{person}{Jack~W. Rae}, \bibinfo{person}{Oriol Vinyals}, {and} \bibinfo{person}{Laurent Sifre}.} \bibinfo{year}{2022}\natexlab{}.
\newblock \showarticletitle{Training Compute-Optimal Large Language Models}.
\newblock \bibinfo{journal}{\emph{arXiv preprint arXiv:2203.15556}} (\bibinfo{year}{2022}).
\newblock


\bibitem[Hu et~al\mbox{.}(2022)]%
        {ref:lora_finetune}
\bibfield{author}{\bibinfo{person}{Edward~J. Hu}, \bibinfo{person}{Yelong Shen}, \bibinfo{person}{Phillip Wallis}, \bibinfo{person}{Zeyuan Allen{-}Zhu}, \bibinfo{person}{Yuanzhi Li}, \bibinfo{person}{Shean Wang}, \bibinfo{person}{Lu Wang}, {and} \bibinfo{person}{Weizhu Chen}.} \bibinfo{year}{2022}\natexlab{}.
\newblock \showarticletitle{LoRA: Low-Rank Adaptation of Large Language Models}. In \bibinfo{booktitle}{\emph{{ICLR}}}. \bibinfo{publisher}{OpenReview.net}.
\newblock


\bibitem[Huang et~al\mbox{.}(2022)]%
        {huang2022large}
\bibfield{author}{\bibinfo{person}{Jiaxin Huang}, \bibinfo{person}{Shixiang~Shane Gu}, \bibinfo{person}{Le Hou}, \bibinfo{person}{Yuexin Wu}, \bibinfo{person}{Xuezhi Wang}, \bibinfo{person}{Hongkun Yu}, {and} \bibinfo{person}{Jiawei Han}.} \bibinfo{year}{2022}\natexlab{}.
\newblock \showarticletitle{Large language models can self-improve}.
\newblock \bibinfo{journal}{\emph{arXiv preprint arXiv:2210.11610}} (\bibinfo{year}{2022}).
\newblock


\bibitem[Jiang et~al\mbox{.}(2021)]%
        {jiang2021can}
\bibfield{author}{\bibinfo{person}{Zhengbao Jiang}, \bibinfo{person}{Jun Araki}, \bibinfo{person}{Haibo Ding}, {and} \bibinfo{person}{Graham Neubig}.} \bibinfo{year}{2021}\natexlab{}.
\newblock \showarticletitle{How can we know when language models know? on the calibration of language models for question answering}.
\newblock \bibinfo{journal}{\emph{Transactions of the Association for Computational Linguistics}}  \bibinfo{volume}{9} (\bibinfo{year}{2021}), \bibinfo{pages}{962--977}.
\newblock


\bibitem[Kadavath et~al\mbox{.}(2022)]%
        {kadavath2022language}
\bibfield{author}{\bibinfo{person}{Saurav Kadavath}, \bibinfo{person}{Tom Conerly}, \bibinfo{person}{Amanda Askell}, \bibinfo{person}{Tom Henighan}, \bibinfo{person}{Dawn Drain}, \bibinfo{person}{Ethan Perez}, \bibinfo{person}{Nicholas Schiefer}, \bibinfo{person}{Zac Hatfield-Dodds}, \bibinfo{person}{Nova DasSarma}, \bibinfo{person}{Eli Tran-Johnson}, {et~al\mbox{.}}} \bibinfo{year}{2022}\natexlab{}.
\newblock \showarticletitle{Language models (mostly) know what they know}.
\newblock \bibinfo{journal}{\emph{arXiv preprint arXiv:2207.05221}} (\bibinfo{year}{2022}).
\newblock


\bibitem[Kocmi and Federmann(2023)]%
        {kocmi2023gemba}
\bibfield{author}{\bibinfo{person}{Tom Kocmi} {and} \bibinfo{person}{Christian Federmann}.} \bibinfo{year}{2023}\natexlab{}.
\newblock \showarticletitle{GEMBA-MQM: Detecting translation quality error spans with GPT-4}.
\newblock \bibinfo{journal}{\emph{arXiv preprint arXiv:2310.13988}} (\bibinfo{year}{2023}).
\newblock


\bibitem[Kojima et~al\mbox{.}(2022)]%
        {kojima2022large}
\bibfield{author}{\bibinfo{person}{Takeshi Kojima}, \bibinfo{person}{Shixiang~Shane Gu}, \bibinfo{person}{Machel Reid}, \bibinfo{person}{Yutaka Matsuo}, {and} \bibinfo{person}{Yusuke Iwasawa}.} \bibinfo{year}{2022}\natexlab{}.
\newblock \showarticletitle{Large Language Models are Zero-Shot Reasoners}.
\newblock \bibinfo{journal}{\emph{arXiv preprint arXiv:2205.11916}} (\bibinfo{year}{2022}).
\newblock


\bibitem[Koo et~al\mbox{.}(2023)]%
        {koo2023benchmarking}
\bibfield{author}{\bibinfo{person}{Ryan Koo}, \bibinfo{person}{Minhwa Lee}, \bibinfo{person}{Vipul Raheja}, \bibinfo{person}{Jong~Inn Park}, \bibinfo{person}{Zae~Myung Kim}, {and} \bibinfo{person}{Dongyeop Kang}.} \bibinfo{year}{2023}\natexlab{}.
\newblock \showarticletitle{Benchmarking cognitive biases in large language models as evaluators}.
\newblock \bibinfo{journal}{\emph{arXiv preprint arXiv:2309.17012}} (\bibinfo{year}{2023}).
\newblock


\bibitem[Kuhn et~al\mbox{.}(2023)]%
        {kuhn2023semantic}
\bibfield{author}{\bibinfo{person}{Lorenz Kuhn}, \bibinfo{person}{Yarin Gal}, {and} \bibinfo{person}{Sebastian Farquhar}.} \bibinfo{year}{2023}\natexlab{}.
\newblock \showarticletitle{Semantic uncertainty: Linguistic invariances for uncertainty estimation in natural language generation}.
\newblock \bibinfo{journal}{\emph{arXiv preprint arXiv:2302.09664}} (\bibinfo{year}{2023}).
\newblock


\bibitem[Li et~al\mbox{.}(2022a)]%
        {li2022making}
\bibfield{author}{\bibinfo{person}{Yifei Li}, \bibinfo{person}{Zeqi Lin}, \bibinfo{person}{Shizhuo Zhang}, \bibinfo{person}{Qiang Fu}, \bibinfo{person}{Bei Chen}, \bibinfo{person}{Jian-Guang Lou}, {and} \bibinfo{person}{Weizhu Chen}.} \bibinfo{year}{2022}\natexlab{a}.
\newblock \showarticletitle{Making large language models better reasoners with step-aware verifier}.
\newblock \bibinfo{journal}{\emph{arXiv preprint arXiv:2206.02336}} (\bibinfo{year}{2022}).
\newblock


\bibitem[Li et~al\mbox{.}(2022b)]%
        {li2022advance}
\bibfield{author}{\bibinfo{person}{Yifei Li}, \bibinfo{person}{Zeqi Lin}, \bibinfo{person}{Shizhuo Zhang}, \bibinfo{person}{Qiang Fu}, \bibinfo{person}{Bei Chen}, \bibinfo{person}{Jian-Guang Lou}, {and} \bibinfo{person}{Weizhu Chen}.} \bibinfo{year}{2022}\natexlab{b}.
\newblock \showarticletitle{On the Advance of Making Language Models Better Reasoners}.
\newblock \bibinfo{journal}{\emph{arXiv preprint arXiv:2206.02336}} (\bibinfo{year}{2022}).
\newblock


\bibitem[Lin et~al\mbox{.}(2023)]%
        {lin2023sphinx}
\bibfield{author}{\bibinfo{person}{Ziyi Lin}, \bibinfo{person}{Chris Liu}, \bibinfo{person}{Renrui Zhang}, \bibinfo{person}{Peng Gao}, \bibinfo{person}{Longtian Qiu}, \bibinfo{person}{Han Xiao}, \bibinfo{person}{Han Qiu}, \bibinfo{person}{Chen Lin}, \bibinfo{person}{Wenqi Shao}, \bibinfo{person}{Keqin Chen}, {et~al\mbox{.}}} \bibinfo{year}{2023}\natexlab{}.
\newblock \showarticletitle{Sphinx: The joint mixing of weights, tasks, and visual embeddings for multi-modal large language models}.
\newblock \bibinfo{journal}{\emph{arXiv preprint arXiv:2311.07575}} (\bibinfo{year}{2023}).
\newblock


\bibitem[Liu et~al\mbox{.}(2024a)]%
        {ref:llava1.5}
\bibfield{author}{\bibinfo{person}{Haotian Liu}, \bibinfo{person}{Chunyuan Li}, \bibinfo{person}{Yuheng Li}, {and} \bibinfo{person}{Yong~Jae Lee}.} \bibinfo{year}{2024}\natexlab{a}.
\newblock \showarticletitle{Improved Baselines with Visual Instruction Tuning}. In \bibinfo{booktitle}{\emph{{CVPR}}}. \bibinfo{publisher}{{IEEE}}, \bibinfo{pages}{26286--26296}.
\newblock


\bibitem[Liu et~al\mbox{.}(2024b)]%
        {liu2024improved}
\bibfield{author}{\bibinfo{person}{Haotian Liu}, \bibinfo{person}{Chunyuan Li}, \bibinfo{person}{Yuheng Li}, {and} \bibinfo{person}{Yong~Jae Lee}.} \bibinfo{year}{2024}\natexlab{b}.
\newblock \showarticletitle{Improved baselines with visual instruction tuning}. In \bibinfo{booktitle}{\emph{Proceedings of the IEEE/CVF Conference on Computer Vision and Pattern Recognition}}. \bibinfo{pages}{26296--26306}.
\newblock


\bibitem[Liu et~al\mbox{.}(2023a)]%
        {liu2023g}
\bibfield{author}{\bibinfo{person}{Yang Liu}, \bibinfo{person}{Dan Iter}, \bibinfo{person}{Yichong Xu}, \bibinfo{person}{Shuohang Wang}, \bibinfo{person}{Ruochen Xu}, {and} \bibinfo{person}{Chenguang Zhu}.} \bibinfo{year}{2023}\natexlab{a}.
\newblock \showarticletitle{G-eval: Nlg evaluation using gpt-4 with better human alignment}.
\newblock \bibinfo{journal}{\emph{arXiv preprint arXiv:2303.16634}} (\bibinfo{year}{2023}).
\newblock


\bibitem[Liu et~al\mbox{.}(2023b)]%
        {liu2023llms}
\bibfield{author}{\bibinfo{person}{Yiqi Liu}, \bibinfo{person}{Nafise~Sadat Moosavi}, {and} \bibinfo{person}{Chenghua Lin}.} \bibinfo{year}{2023}\natexlab{b}.
\newblock \showarticletitle{Llms as narcissistic evaluators: When ego inflates evaluation scores}.
\newblock \bibinfo{journal}{\emph{arXiv preprint arXiv:2311.09766}} (\bibinfo{year}{2023}).
\newblock


\bibitem[Lu et~al\mbox{.}(2024)]%
        {lu2024deepseek}
\bibfield{author}{\bibinfo{person}{Haoyu Lu}, \bibinfo{person}{Wen Liu}, \bibinfo{person}{Bo Zhang}, \bibinfo{person}{Bingxuan Wang}, \bibinfo{person}{Kai Dong}, \bibinfo{person}{Bo Liu}, \bibinfo{person}{Jingxiang Sun}, \bibinfo{person}{Tongzheng Ren}, \bibinfo{person}{Zhuoshu Li}, \bibinfo{person}{Hao Yang}, {et~al\mbox{.}}} \bibinfo{year}{2024}\natexlab{}.
\newblock \showarticletitle{Deepseek-vl: towards real-world vision-language understanding}.
\newblock \bibinfo{journal}{\emph{arXiv preprint arXiv:2403.05525}} (\bibinfo{year}{2024}).
\newblock


\bibitem[Lu et~al\mbox{.}(2023)]%
        {mathvista}
\bibfield{author}{\bibinfo{person}{Pan Lu}, \bibinfo{person}{Hritik Bansal}, \bibinfo{person}{Tony Xia}, \bibinfo{person}{Jiacheng Liu}, \bibinfo{person}{Chunyuan Li}, \bibinfo{person}{Hannaneh Hajishirzi}, \bibinfo{person}{Hao Cheng}, \bibinfo{person}{Kai-Wei Chang}, \bibinfo{person}{Michel Galley}, {and} \bibinfo{person}{Jianfeng Gao}.} \bibinfo{year}{2023}\natexlab{}.
\newblock \showarticletitle{Mathvista: Evaluating mathematical reasoning of foundation models in visual contexts}.
\newblock \bibinfo{journal}{\emph{arXiv preprint arXiv:2310.02255}} (\bibinfo{year}{2023}).
\newblock


\bibitem[Madaan et~al\mbox{.}(2024)]%
        {madaan2024self}
\bibfield{author}{\bibinfo{person}{Aman Madaan}, \bibinfo{person}{Niket Tandon}, \bibinfo{person}{Prakhar Gupta}, \bibinfo{person}{Skyler Hallinan}, \bibinfo{person}{Luyu Gao}, \bibinfo{person}{Sarah Wiegreffe}, \bibinfo{person}{Uri Alon}, \bibinfo{person}{Nouha Dziri}, \bibinfo{person}{Shrimai Prabhumoye}, \bibinfo{person}{Yiming Yang}, {et~al\mbox{.}}} \bibinfo{year}{2024}\natexlab{}.
\newblock \showarticletitle{Self-refine: Iterative refinement with self-feedback}.
\newblock \bibinfo{journal}{\emph{Advances in Neural Information Processing Systems}}  \bibinfo{volume}{36} (\bibinfo{year}{2024}).
\newblock


\bibitem[OpenAI(2023)]%
        {openai20234v}
\bibfield{author}{\bibinfo{person}{GPT OpenAI}.} \bibinfo{year}{2023}\natexlab{}.
\newblock \showarticletitle{4V (ision) system card}.
\newblock \bibinfo{journal}{\emph{preprint}} (\bibinfo{year}{2023}).
\newblock


\bibitem[Pan et~al\mbox{.}(2023)]%
        {pan2023plum}
\bibfield{author}{\bibinfo{person}{Rui Pan}, \bibinfo{person}{Shuo Xing}, \bibinfo{person}{Shizhe Diao}, \bibinfo{person}{Xiang Liu}, \bibinfo{person}{Kashun Shum}, \bibinfo{person}{Jipeng Zhang}, {and} \bibinfo{person}{Tong Zhang}.} \bibinfo{year}{2023}\natexlab{}.
\newblock \showarticletitle{Plum: Prompt learning using metaheuristic}.
\newblock \bibinfo{journal}{\emph{arXiv preprint arXiv:2311.08364}} (\bibinfo{year}{2023}).
\newblock


\bibitem[Pan et~al\mbox{.}(2024)]%
        {pan2024pomp}
\bibfield{author}{\bibinfo{person}{Shilong Pan}, \bibinfo{person}{Zhiliang Tian}, \bibinfo{person}{Liang Ding}, \bibinfo{person}{Zhen Huang}, \bibinfo{person}{Zhihua Wen}, {and} \bibinfo{person}{Dongsheng Li}.} \bibinfo{year}{2024}\natexlab{}.
\newblock \bibinfo{title}{POMP: Probability-driven Meta-graph Prompter for LLMs in Low-resource Unsupervised Neural Machine Translation}.
\newblock
\showeprint[arxiv]{2401.05596}~[cs.CL]


\bibitem[Paul et~al\mbox{.}(2023)]%
        {paul2023refiner}
\bibfield{author}{\bibinfo{person}{Debjit Paul}, \bibinfo{person}{Mete Ismayilzada}, \bibinfo{person}{Maxime Peyrard}, \bibinfo{person}{Beatriz Borges}, \bibinfo{person}{Antoine Bosselut}, \bibinfo{person}{Robert West}, {and} \bibinfo{person}{Boi Faltings}.} \bibinfo{year}{2023}\natexlab{}.
\newblock \showarticletitle{Refiner: Reasoning feedback on intermediate representations}.
\newblock \bibinfo{journal}{\emph{arXiv preprint arXiv:2304.01904}} (\bibinfo{year}{2023}).
\newblock


\bibitem[Qiao et~al\mbox{.}(2024)]%
        {wemath}
\bibfield{author}{\bibinfo{person}{Runqi Qiao}, \bibinfo{person}{Qiuna Tan}, \bibinfo{person}{Guanting Dong}, \bibinfo{person}{Minhui Wu}, \bibinfo{person}{Chong Sun}, \bibinfo{person}{Xiaoshuai Song}, \bibinfo{person}{Zhuoma GongQue}, \bibinfo{person}{Shanglin Lei}, \bibinfo{person}{Zhe Wei}, \bibinfo{person}{Miaoxuan Zhang}, {et~al\mbox{.}}} \bibinfo{year}{2024}\natexlab{}.
\newblock \showarticletitle{We-math: Does your large multimodal model achieve human-like mathematical reasoning?}
\newblock \bibinfo{journal}{\emph{arXiv preprint arXiv:2407.01284}} (\bibinfo{year}{2024}).
\newblock


\bibitem[Rae et~al\mbox{.}(2021)]%
        {rae2021scaling}
\bibfield{author}{\bibinfo{person}{Jack~W Rae}, \bibinfo{person}{Sebastian Borgeaud}, \bibinfo{person}{Trevor Cai}, \bibinfo{person}{Katie Millican}, \bibinfo{person}{Jordan Hoffmann}, \bibinfo{person}{Francis Song}, \bibinfo{person}{John Aslanides}, \bibinfo{person}{Sarah Henderson}, \bibinfo{person}{Roman Ring}, \bibinfo{person}{Susannah Young}, {et~al\mbox{.}}} \bibinfo{year}{2021}\natexlab{}.
\newblock \showarticletitle{Scaling language models: Methods, analysis \& insights from training gopher}.
\newblock \bibinfo{journal}{\emph{arXiv preprint arXiv:2112.11446}} (\bibinfo{year}{2021}).
\newblock


\bibitem[Shi et~al\mbox{.}(2024)]%
        {mathllava_math360k}
\bibfield{author}{\bibinfo{person}{Wenhao Shi}, \bibinfo{person}{Zhiqiang Hu}, \bibinfo{person}{Yi Bin}, \bibinfo{person}{Junhua Liu}, \bibinfo{person}{Yang Yang}, \bibinfo{person}{See-Kiong Ng}, \bibinfo{person}{Lidong Bing}, {and} \bibinfo{person}{Roy Ka-Wei Lee}.} \bibinfo{year}{2024}\natexlab{}.
\newblock \showarticletitle{Math-llava: Bootstrapping mathematical reasoning for multimodal large language models}.
\newblock \bibinfo{journal}{\emph{arXiv preprint arXiv:2406.17294}} (\bibinfo{year}{2024}).
\newblock


\bibitem[Shinn et~al\mbox{.}(2023)]%
        {shinn2023reflexion}
\bibfield{author}{\bibinfo{person}{Noah Shinn}, \bibinfo{person}{Beck Labash}, {and} \bibinfo{person}{Ashwin Gopinath}.} \bibinfo{year}{2023}\natexlab{}.
\newblock \showarticletitle{Reflexion: an autonomous agent with dynamic memory and self-reflection}.
\newblock \bibinfo{journal}{\emph{arXiv preprint arXiv:2303.11366}} \bibinfo{volume}{2}, \bibinfo{number}{5} (\bibinfo{year}{2023}), \bibinfo{pages}{9}.
\newblock


\bibitem[Shum et~al\mbox{.}(2023)]%
        {shum2023automatic}
\bibfield{author}{\bibinfo{person}{Kashun Shum}, \bibinfo{person}{Shizhe Diao}, {and} \bibinfo{person}{Tong Zhang}.} \bibinfo{year}{2023}\natexlab{}.
\newblock \showarticletitle{Automatic Prompt Augmentation and Selection with Chain-of-Thought from Labeled Data}. In \bibinfo{booktitle}{\emph{Findings of the Association for Computational Linguistics: EMNLP 2023}}. \bibinfo{pages}{12113--12139}.
\newblock


\bibitem[Team et~al\mbox{.}(2023)]%
        {team2023gemini}
\bibfield{author}{\bibinfo{person}{Gemini Team}, \bibinfo{person}{Rohan Anil}, \bibinfo{person}{Sebastian Borgeaud}, \bibinfo{person}{Jean-Baptiste Alayrac}, \bibinfo{person}{Jiahui Yu}, \bibinfo{person}{Radu Soricut}, \bibinfo{person}{Johan Schalkwyk}, \bibinfo{person}{Andrew~M Dai}, \bibinfo{person}{Anja Hauth}, \bibinfo{person}{Katie Millican}, {et~al\mbox{.}}} \bibinfo{year}{2023}\natexlab{}.
\newblock \showarticletitle{Gemini: a family of highly capable multimodal models}.
\newblock \bibinfo{journal}{\emph{arXiv preprint arXiv:2312.11805}} (\bibinfo{year}{2023}).
\newblock


\bibitem[Wadhwa et~al\mbox{.}(2024)]%
        {ref:cot_distill_2}
\bibfield{author}{\bibinfo{person}{Somin Wadhwa}, \bibinfo{person}{Silvio Amir}, {and} \bibinfo{person}{Byron~C. Wallace}.} \bibinfo{year}{2024}\natexlab{}.
\newblock \showarticletitle{Investigating Mysteries of CoT-Augmented Distillation}. In \bibinfo{booktitle}{\emph{{EMNLP}}}. \bibinfo{publisher}{Association for Computational Linguistics}, \bibinfo{pages}{6071--6086}.
\newblock


\bibitem[Wang et~al\mbox{.}(2024b)]%
        {math-v}
\bibfield{author}{\bibinfo{person}{Ke Wang}, \bibinfo{person}{Junting Pan}, \bibinfo{person}{Weikang Shi}, \bibinfo{person}{Zimu Lu}, \bibinfo{person}{Mingjie Zhan}, {and} \bibinfo{person}{Hongsheng Li}.} \bibinfo{year}{2024}\natexlab{b}.
\newblock \showarticletitle{Measuring multimodal mathematical reasoning with math-vision dataset}.
\newblock \bibinfo{journal}{\emph{arXiv preprint arXiv:2402.14804}} (\bibinfo{year}{2024}).
\newblock


\bibitem[Wang et~al\mbox{.}(2024a)]%
        {ref:qwen2-vl}
\bibfield{author}{\bibinfo{person}{Peng Wang}, \bibinfo{person}{Shuai Bai}, \bibinfo{person}{Sinan Tan}, \bibinfo{person}{Shijie Wang}, \bibinfo{person}{Zhihao Fan}, \bibinfo{person}{Jinze Bai}, \bibinfo{person}{Keqin Chen}, \bibinfo{person}{Xuejing Liu}, \bibinfo{person}{Jialin Wang}, \bibinfo{person}{Wenbin Ge}, \bibinfo{person}{Yang Fan}, \bibinfo{person}{Kai Dang}, \bibinfo{person}{Mengfei Du}, \bibinfo{person}{Xuancheng Ren}, \bibinfo{person}{Rui Men}, \bibinfo{person}{Dayiheng Liu}, \bibinfo{person}{Chang Zhou}, \bibinfo{person}{Jingren Zhou}, {and} \bibinfo{person}{Junyang Lin}.} \bibinfo{year}{2024}\natexlab{a}.
\newblock \showarticletitle{Qwen2-VL: Enhancing Vision-Language Model's Perception of the World at Any Resolution}.
\newblock \bibinfo{journal}{\emph{CoRR}}  \bibinfo{volume}{abs/2409.12191} (\bibinfo{year}{2024}).
\newblock


\bibitem[Wang et~al\mbox{.}(2023b)]%
        {ref:cot_distill_1}
\bibfield{author}{\bibinfo{person}{Peifeng Wang}, \bibinfo{person}{Zhengyang Wang}, \bibinfo{person}{Zheng Li}, \bibinfo{person}{Yifan Gao}, \bibinfo{person}{Bing Yin}, {and} \bibinfo{person}{Xiang Ren}.} \bibinfo{year}{2023}\natexlab{b}.
\newblock \showarticletitle{{SCOTT:} Self-Consistent Chain-of-Thought Distillation}. In \bibinfo{booktitle}{\emph{{ACL} {(1)}}}. \bibinfo{publisher}{Association for Computational Linguistics}, \bibinfo{pages}{5546--5558}.
\newblock


\bibitem[Wang et~al\mbox{.}(2023a)]%
        {wang2023cogvlm}
\bibfield{author}{\bibinfo{person}{Weihan Wang}, \bibinfo{person}{Qingsong Lv}, \bibinfo{person}{Wenmeng Yu}, \bibinfo{person}{Wenyi Hong}, \bibinfo{person}{Ji Qi}, \bibinfo{person}{Yan Wang}, \bibinfo{person}{Junhui Ji}, \bibinfo{person}{Zhuoyi Yang}, \bibinfo{person}{Lei Zhao}, \bibinfo{person}{Xixuan Song}, {et~al\mbox{.}}} \bibinfo{year}{2023}\natexlab{a}.
\newblock \showarticletitle{Cogvlm: Visual expert for pretrained language models}.
\newblock \bibinfo{journal}{\emph{arXiv preprint arXiv:2311.03079}} (\bibinfo{year}{2023}).
\newblock


\bibitem[Wang et~al\mbox{.}(2022a)]%
        {wang2022rationale}
\bibfield{author}{\bibinfo{person}{Xuezhi Wang}, \bibinfo{person}{Jason Wei}, \bibinfo{person}{Dale Schuurmans}, \bibinfo{person}{Quoc Le}, \bibinfo{person}{Ed Chi}, {and} \bibinfo{person}{Denny Zhou}.} \bibinfo{year}{2022}\natexlab{a}.
\newblock \showarticletitle{Rationale-Augmented Ensembles in Language Models}.
\newblock \bibinfo{journal}{\emph{arXiv preprint arXiv:2207.00747}} (\bibinfo{year}{2022}).
\newblock


\bibitem[Wang et~al\mbox{.}(2022b)]%
        {wang2022self}
\bibfield{author}{\bibinfo{person}{Xuezhi Wang}, \bibinfo{person}{Jason Wei}, \bibinfo{person}{Dale Schuurmans}, \bibinfo{person}{Quoc Le}, \bibinfo{person}{Ed Chi}, {and} \bibinfo{person}{Denny Zhou}.} \bibinfo{year}{2022}\natexlab{b}.
\newblock \showarticletitle{Self-consistency improves chain of thought reasoning in language models}.
\newblock \bibinfo{journal}{\emph{arXiv preprint arXiv:2203.11171}} (\bibinfo{year}{2022}).
\newblock


\bibitem[Wei et~al\mbox{.}(2022a)]%
        {wei2022emergent}
\bibfield{author}{\bibinfo{person}{Jason Wei}, \bibinfo{person}{Yi Tay}, \bibinfo{person}{Rishi Bommasani}, \bibinfo{person}{Colin Raffel}, \bibinfo{person}{Barret Zoph}, \bibinfo{person}{Sebastian Borgeaud}, \bibinfo{person}{Dani Yogatama}, \bibinfo{person}{Maarten Bosma}, \bibinfo{person}{Denny Zhou}, \bibinfo{person}{Donald Metzler}, {et~al\mbox{.}}} \bibinfo{year}{2022}\natexlab{a}.
\newblock \showarticletitle{Emergent abilities of large language models}.
\newblock \bibinfo{journal}{\emph{arXiv preprint arXiv:2206.07682}} (\bibinfo{year}{2022}).
\newblock


\bibitem[Wei et~al\mbox{.}(2022b)]%
        {wei2022chain}
\bibfield{author}{\bibinfo{person}{Jason Wei}, \bibinfo{person}{Xuezhi Wang}, \bibinfo{person}{Dale Schuurmans}, \bibinfo{person}{Maarten Bosma}, \bibinfo{person}{Ed Chi}, \bibinfo{person}{Quoc Le}, {and} \bibinfo{person}{Denny Zhou}.} \bibinfo{year}{2022}\natexlab{b}.
\newblock \showarticletitle{Chain of thought prompting elicits reasoning in large language models}.
\newblock \bibinfo{journal}{\emph{arXiv preprint arXiv:2201.11903}} (\bibinfo{year}{2022}).
\newblock


\bibitem[Wei et~al\mbox{.}(2022c)]%
        {ref:CoT}
\bibfield{author}{\bibinfo{person}{Jason Wei}, \bibinfo{person}{Xuezhi Wang}, \bibinfo{person}{Dale Schuurmans}, \bibinfo{person}{Maarten Bosma}, \bibinfo{person}{Fei Xia}, \bibinfo{person}{Ed Chi}, \bibinfo{person}{Quoc~V Le}, \bibinfo{person}{Denny Zhou}, {et~al\mbox{.}}} \bibinfo{year}{2022}\natexlab{c}.
\newblock \showarticletitle{Chain-of-thought prompting elicits reasoning in large language models}.
\newblock \bibinfo{journal}{\emph{Advances in neural information processing systems}}  \bibinfo{volume}{35} (\bibinfo{year}{2022}), \bibinfo{pages}{24824--24837}.
\newblock


\bibitem[Xiong et~al\mbox{.}(2024)]%
        {ref:self_evaluation}
\bibfield{author}{\bibinfo{person}{Tianyi Xiong}, \bibinfo{person}{Xiyao Wang}, \bibinfo{person}{Dong Guo}, \bibinfo{person}{Qinghao Ye}, \bibinfo{person}{Haoqi Fan}, \bibinfo{person}{Quanquan Gu}, \bibinfo{person}{Heng Huang}, {and} \bibinfo{person}{Chunyuan Li}.} \bibinfo{year}{2024}\natexlab{}.
\newblock \showarticletitle{Llava-critic: Learning to evaluate multimodal models}.
\newblock \bibinfo{journal}{\emph{arXiv preprint arXiv:2410.02712}} (\bibinfo{year}{2024}).
\newblock


\bibitem[Xu et~al\mbox{.}(2023)]%
        {xu2023instructscore}
\bibfield{author}{\bibinfo{person}{Wenda Xu}, \bibinfo{person}{Danqing Wang}, \bibinfo{person}{Liangming Pan}, \bibinfo{person}{Zhenqiao Song}, \bibinfo{person}{Markus Freitag}, \bibinfo{person}{William~Yang Wang}, {and} \bibinfo{person}{Lei Li}.} \bibinfo{year}{2023}\natexlab{}.
\newblock \showarticletitle{INSTRUCTSCORE: Explainable Text Generation Evaluation with Finegrained Feedback}.
\newblock \bibinfo{journal}{\emph{arXiv preprint arXiv:2305.14282}} (\bibinfo{year}{2023}).
\newblock


\bibitem[Xu et~al\mbox{.}(2024)]%
        {xu2024can}
\bibfield{author}{\bibinfo{person}{Xin Xu}, \bibinfo{person}{Shizhe Diao}, \bibinfo{person}{Can Yang}, {and} \bibinfo{person}{Yang Wang}.} \bibinfo{year}{2024}\natexlab{}.
\newblock \showarticletitle{Can We Verify Step by Step for Incorrect Answer Detection?}
\newblock \bibinfo{journal}{\emph{arXiv preprint arXiv:2402.10528}} (\bibinfo{year}{2024}).
\newblock


\bibitem[Yue et~al\mbox{.}(2024)]%
        {yue2024mmmu}
\bibfield{author}{\bibinfo{person}{Xiang Yue}, \bibinfo{person}{Yuansheng Ni}, \bibinfo{person}{Kai Zhang}, \bibinfo{person}{Tianyu Zheng}, \bibinfo{person}{Ruoqi Liu}, \bibinfo{person}{Ge Zhang}, \bibinfo{person}{Samuel Stevens}, \bibinfo{person}{Dongfu Jiang}, \bibinfo{person}{Weiming Ren}, \bibinfo{person}{Yuxuan Sun}, {et~al\mbox{.}}} \bibinfo{year}{2024}\natexlab{}.
\newblock \showarticletitle{Mmmu: A massive multi-discipline multimodal understanding and reasoning benchmark for expert agi}. In \bibinfo{booktitle}{\emph{Proceedings of the IEEE/CVF Conference on Computer Vision and Pattern Recognition}}. \bibinfo{pages}{9556--9567}.
\newblock


\bibitem[Zelikman et~al\mbox{.}(2022)]%
        {zelikman2022star}
\bibfield{author}{\bibinfo{person}{Eric Zelikman}, \bibinfo{person}{Yuhuai Wu}, \bibinfo{person}{Jesse Mu}, {and} \bibinfo{person}{Noah Goodman}.} \bibinfo{year}{2022}\natexlab{}.
\newblock \showarticletitle{Star: Bootstrapping reasoning with reasoning}.
\newblock \bibinfo{journal}{\emph{Advances in Neural Information Processing Systems}}  \bibinfo{volume}{35} (\bibinfo{year}{2022}), \bibinfo{pages}{15476--15488}.
\newblock


\bibitem[Zhang et~al\mbox{.}(2023)]%
        {zhang2023coder}
\bibfield{author}{\bibinfo{person}{Tianyi Zhang}, \bibinfo{person}{Tao Yu}, \bibinfo{person}{Tatsunori Hashimoto}, \bibinfo{person}{Mike Lewis}, \bibinfo{person}{Wen-tau Yih}, \bibinfo{person}{Daniel Fried}, {and} \bibinfo{person}{Sida Wang}.} \bibinfo{year}{2023}\natexlab{}.
\newblock \showarticletitle{Coder reviewer reranking for code generation}. In \bibinfo{booktitle}{\emph{International Conference on Machine Learning}}. PMLR, \bibinfo{pages}{41832--41846}.
\newblock


\bibitem[Zhang et~al\mbox{.}(2022)]%
        {zhang2022automatic}
\bibfield{author}{\bibinfo{person}{Zhuosheng Zhang}, \bibinfo{person}{Aston Zhang}, \bibinfo{person}{Mu Li}, {and} \bibinfo{person}{Alex Smola}.} \bibinfo{year}{2022}\natexlab{}.
\newblock \showarticletitle{Automatic chain of thought prompting in large language models}.
\newblock \bibinfo{journal}{\emph{arXiv preprint arXiv:2210.03493}} (\bibinfo{year}{2022}).
\newblock


\bibitem[Zheng et~al\mbox{.}(2023)]%
        {zheng2023judging}
\bibfield{author}{\bibinfo{person}{Lianmin Zheng}, \bibinfo{person}{Wei-Lin Chiang}, \bibinfo{person}{Ying Sheng}, \bibinfo{person}{Siyuan Zhuang}, \bibinfo{person}{Zhanghao Wu}, \bibinfo{person}{Yonghao Zhuang}, \bibinfo{person}{Zi Lin}, \bibinfo{person}{Zhuohan Li}, \bibinfo{person}{Dacheng Li}, \bibinfo{person}{Eric Xing}, {et~al\mbox{.}}} \bibinfo{year}{2023}\natexlab{}.
\newblock \showarticletitle{Judging llm-as-a-judge with mt-bench and chatbot arena}.
\newblock \bibinfo{journal}{\emph{Advances in Neural Information Processing Systems}}  \bibinfo{volume}{36} (\bibinfo{year}{2023}), \bibinfo{pages}{46595--46623}.
\newblock


\bibitem[Zhou et~al\mbox{.}(2022)]%
        {zhou2022least}
\bibfield{author}{\bibinfo{person}{Denny Zhou}, \bibinfo{person}{Nathanael Sch{\"a}rli}, \bibinfo{person}{Le Hou}, \bibinfo{person}{Jason Wei}, \bibinfo{person}{Nathan Scales}, \bibinfo{person}{Xuezhi Wang}, \bibinfo{person}{Dale Schuurmans}, \bibinfo{person}{Olivier Bousquet}, \bibinfo{person}{Quoc Le}, {and} \bibinfo{person}{Ed Chi}.} \bibinfo{year}{2022}\natexlab{}.
\newblock \showarticletitle{Least-to-Most Prompting Enables Complex Reasoning in Large Language Models}.
\newblock \bibinfo{journal}{\emph{arXiv preprint arXiv:2205.10625}} (\bibinfo{year}{2022}).
\newblock


\bibitem[Zhu et~al\mbox{.}(2023)]%
        {zhu2023minigpt}
\bibfield{author}{\bibinfo{person}{Deyao Zhu}, \bibinfo{person}{Jun Chen}, \bibinfo{person}{Xiaoqian Shen}, \bibinfo{person}{Xiang Li}, {and} \bibinfo{person}{Mohamed Elhoseiny}.} \bibinfo{year}{2023}\natexlab{}.
\newblock \showarticletitle{Minigpt-4: Enhancing vision-language understanding with advanced large language models}.
\newblock \bibinfo{journal}{\emph{arXiv preprint arXiv:2304.10592}} (\bibinfo{year}{2023}).
\newblock


\end{thebibliography}
